\documentclass[10pt,twocolumn,letterpaper]{article}

\usepackage{iccv}
\usepackage{times}
\usepackage{epsfig}
\usepackage{graphicx}
\usepackage{amsmath}
\usepackage{amssymb}
\usepackage{multirow}
\usepackage{subfigure}
\usepackage[misc]{ifsym}

\makeatletter
\newcommand*\bigcdot{\mathpalette\bigcdot@{.5}}
\newcommand*\bigcdot@[2]{\mathbin{\vcenter{\hbox{\scalebox{#2}{$\m@th#1\bullet$}}}}}
\makeatother

\makeatletter
\renewcommand{\@thesubfigure}{\hskip\subfiglabelskip}
\makeatother

\usepackage[pagebackref=true,breaklinks=true,letterpaper=true,colorlinks,bookmarks=false]{hyperref}

\iccvfinalcopy 

\def\iccvPaperID{6354} 

\ificcvfinal\fi

\begin{document}

\title{Domain Adaptive SiamRPN++ for Object Tracking in the Wild}

\author{Zhongzhou Zhang, Lei Zhang \textsuperscript{(\Letter)} \\
Learning Intelligence $$\&$$ Vision Essential (LiVE) Group \\
School of Microelectronics and Communication Engineering, Chongqing University, China\\
{\tt\small $$\{$$zz.zhang, leizhang$$\}$$@cqu.edu.cn}
}

\maketitle
\ificcvfinal\thispagestyle{empty}\fi

\begin{abstract}
   Benefit from large-scale training data, recent advances in Siamese-based object tracking have achieved compelling results on the normal sequences. Whilst Siamese-based trackers assume training and test data follow an identical distribution. Suppose there is a set of foggy or rainy test sequences, it cannot be guaranteed that the trackers trained on the normal images perform well on the data belonging to other domains. The problem of domain shift among training and test data has already been discussed in object detection and semantic segmentation areas, which, however, has not been investigated for visual tracking. To this end, based on SiamRPN++, we introduce a \textbf{D}omain \textbf{A}daptive \textbf{SiamRPN++}, namely \textbf{DASiamRPN++}, to improve the cross-domain transferability and robustness of a tracker. Inspired by $\mathcal{A}$-distance theory, we present two domain adaptive modules, Pixel Domain Adaptation (PDA) and Semantic Domain Adaptation (SDA). The PDA module aligns the feature maps of template and search region images to eliminate the pixel-level domain shift caused by weather, illumination, etc. The SDA module aligns the feature representations of the tracking target's appearance to eliminate the semantic-level domain shift. PDA and SDA modules reduce the domain disparity by learning domain classifiers in an adversarial training manner. The domain classifiers enforce the network to learn domain-invariant feature representations. Extensive experiments are performed on the standard datasets of two different domains, including synthetic foggy and TIR sequences, which demonstrates the transferability and domain adaptability of the proposed tracker.
\end{abstract}

\section{Introduction}

Visual object tracking is one of the fundamental computer vision problems, which aims to estimate the trajectory of an arbitrary visual target when only an initial state of the target is available. Generic visual tracking is an interesting yet challenging research topic with a wide range of applications, such as video surveillance, autopilot, etc.

\begin{figure}
    \begin{center}
    \includegraphics[scale=0.4]{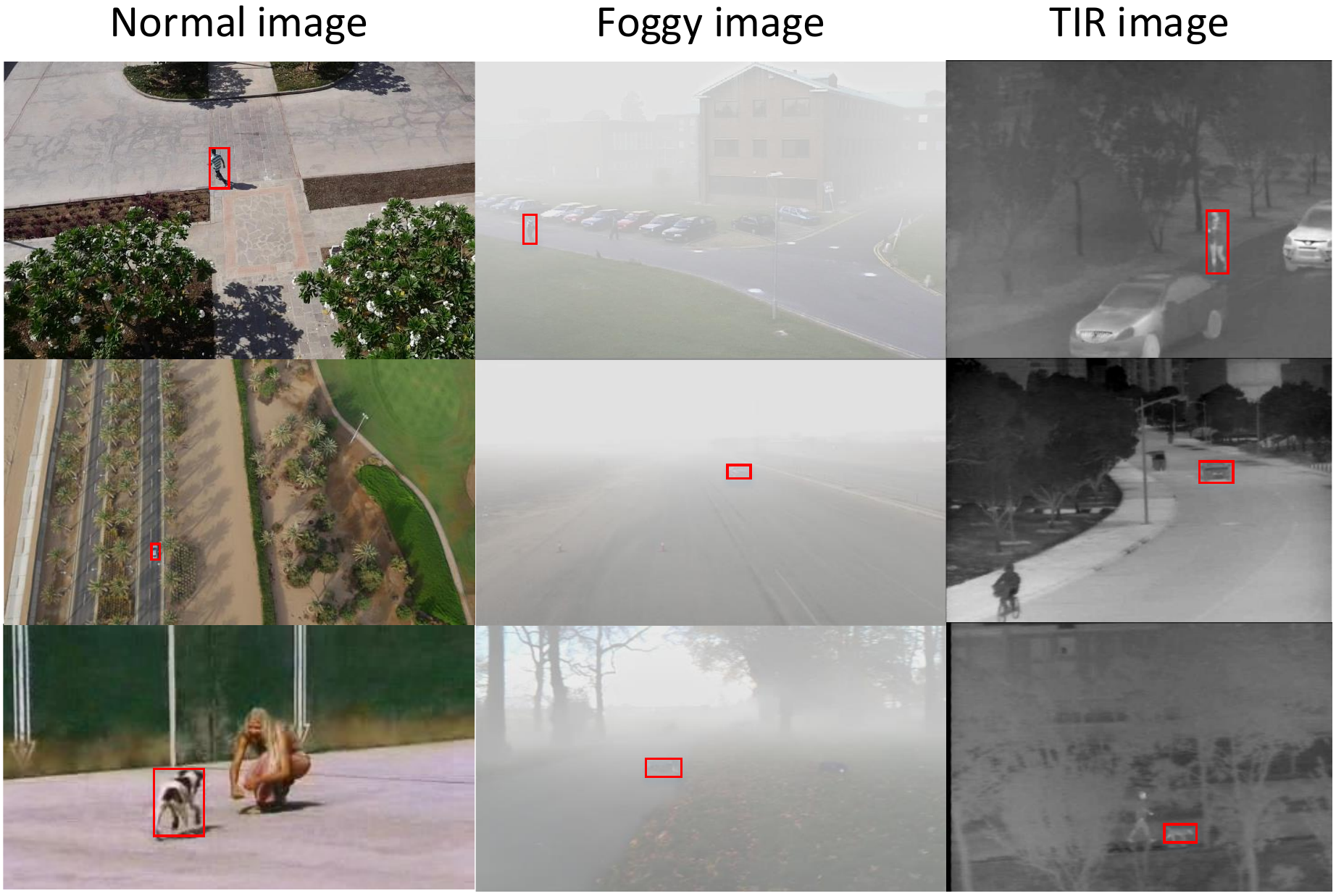}
       \caption{Illustrated images from different domains, following normal images \cite{mueller2016benchmark, 7001050}, foggy images \cite{fan2019lasot, 7001050} and thermal infrared (TIR) images \cite{Kristan2019a}.}
       \vspace{-0.6cm}
    \end{center}
\label{domain}
\end{figure}

Driven by large-scale well-labeled datasets \cite{fan2019lasot, huang2019got, muller2018trackingnet, real2017youtube, russakovsky2015imagenet} and deep convolutional neural networks \cite{he2016deep}, recent object tracking methods have achieved excellent performance, especially for the Siamese-based trackers. The Siamese-based trackers \cite{bertinetto2016fully, fan2019siamese, li2019siamrpn++, li2018high, tao2016siamese, wang2019spm, zhang2019deeper} typically learn a general similarity map by cross-correlation between the feature representations learned from the \emph{template} and \emph{search region}, which are trained with large-scale data in an end-to-end manner. 

\begin{figure}
        \center
        \includegraphics[scale=0.65]{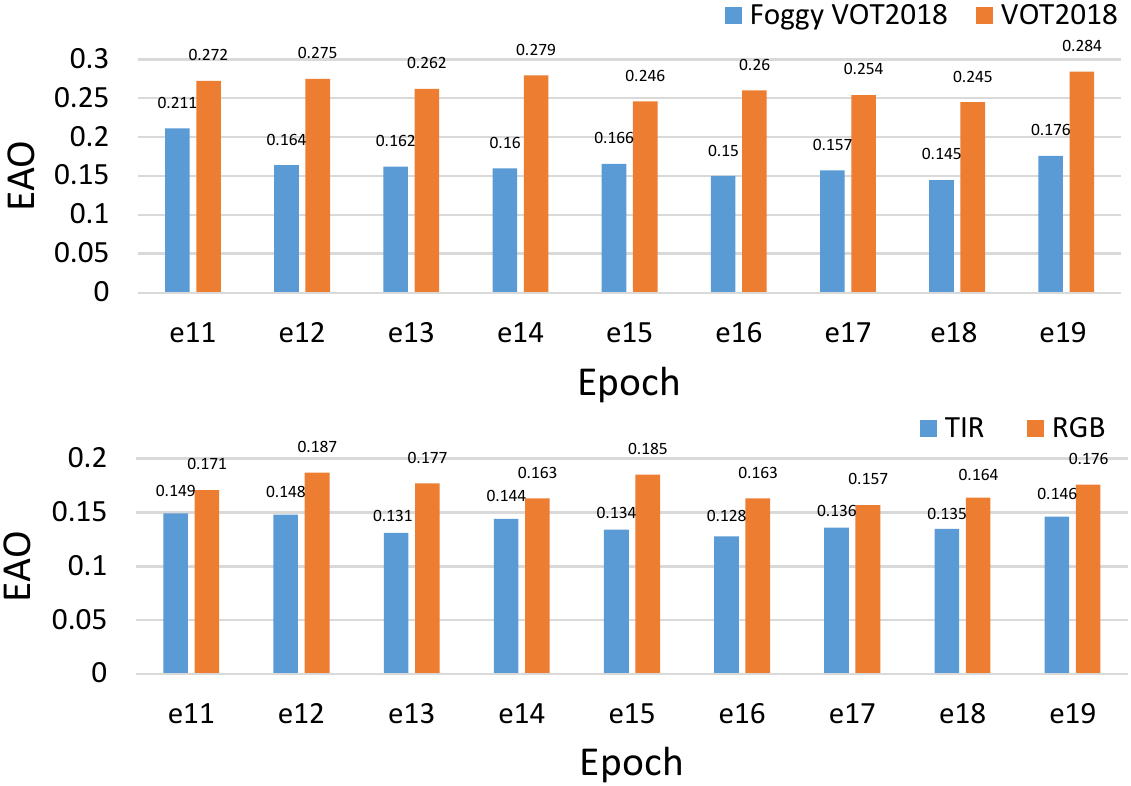}
           \caption{Domain shift causes a performance drop. The figure shows the experimental results on Foggy VOT2018, VOT2018, VOT2019-RGBT (RGB) and VOT2019-RGBT (TIR).}
           \vspace{-0.3cm}
\label{drop}
\end{figure}

\textbf{Problem. }In the real-world scenarios, tracking algorithms are facing enormous challenges due to the diversity of the application environments, such as weather (foggy/rainy/cloudy), modality (RGB/TIR) and illumination (day/night), etc. As illustrated in Fig. \ref{domain}, the sequences, including the tracking targets of \emph{Persons}, \emph{Cars} and \emph{Dogs}, are collected from different domains. In general, the trackers are only trained with the sequences collected in the ideal conditions, and their adaptability and transferability to open scenarios cannot be guaranteed due to domain shift across training and test data. To prove the existence of this problem, we perform the confirmatory experiments by evaluating the pre-trained SiamRPN++ \cite{li2019siamrpn++} on the normal sequences, the thermal infrared (TIR) sequences and the generated foggy sequences. Note that SiamRPN++ is only trained on the LaSOT \cite{fan2019lasot} dataset. (1) We first evaluate the trained models on the synthetic foggy VOT2018 \cite{Kristan2018a} and the normal VOT2018. (2) VOT2019-RGBT \cite{Kristan2019a} benchmark can be split into the RGB sequences and the TIR sequences, and we also evaluate the trained models on the RGB sequences and the TIR sequences, resp. The results are shown in Fig. \ref{drop}, which demonstrate that domain distribution discrepancy between training and test data indeed leads to a significant performance degradation.

One of the solutions to the problem is to collect as much training data as possible to alleviate the impact of domain shift. However, bounding boxes annotation is labor-intensive and time-consuming. Therefore, similar to domain adaptive detection \cite{chen2018domain, he2019multi, he2020domain}, it is reasonable and natural to develop a domain adaptive tracker to handle the performance drop caused by the cross-domain discrepancy. To overcome the cross-domain tracking restriction and improve the generalization ability, we introduce a domain adaptive tracker based on the representative Siamese-based mehtod \cite{li2019siamrpn++}. Following the unsupervised domain adaptation scenario where the source domain is well-labeled while the target domain is unlabeled, we intend to make the domain adaptive tracker perform well on both source and target domain at no additional annotation cost. To the best of our knowledge, we first notice the problem of domain distribution discrepancy in the visual tracking area and propose a domain adaptive tracker.

The general idea in unsupervised domain adaptation is to bridge the domain gap by explicitly learning domain-invariant representations between different domains and achieving small errors on the target domain. We aim to minimize the $\mathcal{A}$-distance \cite{ben2006analysis} which is typically used to measure the distribution divergency between the source and target domain. Based on Bayes's Formula and covariate shift assumption, the probabilistic analysis for tracking is given in Sec \ref{4.2}. Inspired by the $\mathcal{A}$-distance theory and the probabilistic perspective, we put forward a domain adaptive tracker in two levels, i.e. Pixel Domain Adaptation (PDA) and Semantic Domain Adaptation (SDA). The PDA module focuses on the domain shift of each feature pixel, i,e, image style, illumination, etc. And the SDA module pays attention to the whole target with appearance and category change caused by domain shift. PDA and SDA reinforce the convolutional neural networks to learn domain-invariant feature maps and feature representations by training the domain classifier and the Siamese network in an adversarial manner. The contributions of this paper can be summarized as follows:
\begin{itemize}
  \item We first introduce the domain distribution discrepancy problem to the visual tracking community. The confirmatory experiments demonstrate that the domain shift leads to a clear performance drop.
  \item We propose the pixel domain adaptation (PDA) module and the semantic domain adaptation (SDA) module for learning domain-invariant features. PDA performs the domain distribution alignment between the source and target domain in image-level. SDA focuses on the domain distribution alignment w.r.t. the tracking target in semantic-level.
  \item We put forward a new scheme to generate foggy images and construct foggy benchmarks. The single-view depth estimation method, MegaDepth \cite{li2018megadepth}, is adopted to predict the depth maps. Inspired by HazeRD \cite{zhang2017hazerd}, we utilise RGB images and their corresponding depth prediction maps to generate Foggy VOT2018 \cite{Kristan2018a}, Foggy OTB100 \cite{7001050}, Foggy UAV123 \cite{mueller2016benchmark} and Foggy GOT-10k \cite{huang2019got}.
\end{itemize}

\section{Related Work}
\label{sec2}
\textbf{Siamese Network for Tracking. }Siamese network has drawn great attention in the visual tracking area. The pioneering works, including SINT \cite{tao2016siamese} and SiamFC \cite{bertinetto2016fully}, are trained with large-scale image pairs to learn a similarity function in an end-to-end manner. Inspired by Faster R-CNN \cite{ren2015faster}, Li \emph{et al}. \cite{li2018high} integrate the Siamese network with Region Proposal Network (RPN), which takes object tracking as a local one-shot object detection problem. To exploit deeper and wider tracking networks, SiamRPN++ \cite{li2019siamrpn++} replaces the modified AlexNet \cite{krizhevsky2012imagenet} with ResNet-50 \cite{he2016deep} to enrich the extracted feature maps. SPM-tracker \cite{wang2019spm} designs a two-stage network, namely coarse matching stage and fine matching stage, which can achieve high localization precision. Cascade-RPN \cite{fan2019siamese} adopts a multi-stage tracking framework with three RPNs cascaded and leverages feature maps of different levels. Anchor-free trackers \cite{chen2020siamese, guo2020siamcar, zhang2020hard} adopt the per-pixel prediction fashion to get accurate bounding boxes and improve computational efficiency. However, most tracking approaches only generate an appearance model of the target in the first frame, and the appearance model is not updated in the following frames. The robustness of these trackers is severely limited. Bhat \emph{et al}. \cite{bhat2019learning} put forward a target model predictor to online optimize the appearance model, capable of exploiting both target and background information to generate a discriminative target model. However, these SOTA trackers do not consider the domain shift across datasets, which, undoubtedly, weakens their domain adaptability and transferability.

\textbf{Domain Adaptation. }Domain adaptation is widely explored for cross-domain image classification and detection \cite{duan2012domain, he2019multi, hoffman2014lsda, long2018conditional, long2016unsupervised, wang2018lstn}, which aims to transfer knowledge from one domain to another by mitigating the distributional discrepancy. A commonly used approach for domain adaptation is to bridge the domain gap by making the extracted features from different domains indistinguishable. Recently, some methods have achieved tremendous success in unsupervised domain adaptation. Early models minimize the disparity between different domains by measuring the domain divergence through some metrics, such as Maximum Mean Discrepancy (MMD), moment matching, etc. Based on adversarial learning, the recently proposed methods either employ Generative Adversarial Network (GAN) \cite{goodfellow2014generative} or adopt adversarial training with Gradient Reverse Layer (GRL) \cite{ganin2015unsupervised}. Some approaches \cite{bousmalis2017unsupervised, li2017mmd, liu2017unsupervised} motivated by GAN achieve pixel-level adaptation with regard to image-to-image translation techniques. Ganin \emph{et al}. \cite{ganin2015unsupervised} proposed the gradient reverse layer, which reverses the gradient during the backpropagation phase for minimax optimization between the feature representation and domain classifier, an intuitive problem of domain adaptation.

\section{Revisiting of SiamRPN++}
\label{sec3}
SiamRPN++ \cite{li2019siamrpn++} consists of two main components, including the Siamese network for feature extraction and the region proposal network for localization. A \emph{template} \emph{z} and its corresponding \emph{search region} \emph{x} are formulated as an input image pair of SiamRPN++.  Siamese-based trackers typically utilize a two-stream network to respectively extract the feature maps of the \emph{template} and the \emph{search region}. SiamRPN++ adopts a combination of depth-wise cross-correlation and fully convolutional layers to assemble a head module for calculating  classification scores and bounding box regression coordinates (offset), which can be formulated as:
\begin{equation}\label{eq1}
\begin{split}
A^{cls}_{w\times h\times 2k}=\psi_c({\lbrack \varphi(z) \rbrack} _{cls} \star {\lbrack \varphi(x) \rbrack} _{cls}),\\
A^{reg}_{w\times h\times 4k}=\psi_r({\lbrack \varphi(z) \rbrack} _{reg} \star {\lbrack \varphi(x) \rbrack} _{reg})
\end{split}
\end{equation}
where $\varphi$ denotes the Siamese network. $[\, \bigcdot \,]_{cls/reg}$ are the adjustment layers to make the features better fit for different tasks. $\star$ represents cross-correlation operation. $k$ is the number of anchors.
The cross-entropy loss and smooth \emph{L}1 loss are equipped to supervise the binary classification and regression branch, resp.

\section{Theoretical Preliminaries}
\label{sec4}
\subsection{$\mathcal{A}$-distance}
Given a source domain $\mathcal{S}$ and a target domain $\mathcal{T}$, for two sets of samples with domain distributions, we usually use the $\mathcal{A}$-distance \cite{ben2006analysis} to measure the distribution difference between the two sets. Suppose $\mathcal{H}$ to be a hypothesis set of domain classifiers and $h$: $x\rightarrow [0, 1]$, one of the domain classifiers, which aims to predict the source sample $x_{\mathcal{S}}$ to be 0 ({i.e., $h(x_{\mathcal{S}})\rightarrow0$}) and the target sample $x_{\mathcal{T}}$ to be 1 (i.e.,  $h(x_{\mathcal{T}}) \rightarrow 1$). The $\mathcal{A}$-distance, which measures the domain divergence, can be formulated as:
\begin{equation}\label{eq2}
  d_\mathcal{A}(\mathcal{S}, \mathcal{T}) = 2\left( 1 - 2\min_{h\in\mathcal{H}} err\left(h(x)\right) \right)
\end{equation}
where $\mathcal{H}$ consists of a set of domain classifiers and $\mathop{\min}\limits_{h\in\mathcal{H}}err(h(x))$ means the prediction error of an ideal domain classifier. Clearly, a small domain classification error means large domain divergence. Most domain adaptation methods typically minimize the domain discrepancy $d_\mathcal{A}(\mathcal{S}, \mathcal{T})$ to implement the features alignment. Therefore, it is equivalent to maximize the error of the ideal domain classifier and there is,
\begin{equation}\label{eq3}
 \begin{split}
  \min_f{d_{\mathcal{A}}(\mathcal{S}, \mathcal{T})} \Leftrightarrow \min_f 2 \left(1 - 2\min_{h\in\mathcal{H}} err\left(h(x)\right) \right), \\
  \Leftrightarrow \max_f \min_{h\in\mathcal{H}} err\left(h(x)\right) \qquad \quad \quad \ \ \
 \end{split}
\end{equation}
where $f$ is the feature representation of sample $x$ from the Siamese network. Clearly, the problem (\ref{eq3}) is a standard minimax optimization problem between the feature extractor $f$ and the domain classifier $h$. We therefore optimize the above minimax problem in an adversarial training manner. The Gradient Reversed Layer (GRL) \cite{ganin2015unsupervised} is a commonly used method for solving this problem.
\subsection{Probabilistic Analysis for Object Tracker}
\label{4.2}
For visual tracking, given an initial state of the target, trackers aims to predict the position of the target in the following frames, which can be regarded as a posterior probability $P(S, B | Z, X)$ learning problem, where $S$ and $B$ denote the classification score and the predicted bounding box, $Z$ and $X$ represent \emph{template} and \emph{search region}, resp. We utilize $P_{\mathcal{S}}(S, B, Z, X)$ and $P_{\mathcal{T}}(S, B, Z, X)$ to represent the joint distributions of the source and target domain. Owing to the existence of  domain shift, the joint distribution of the source domain is not consistent with that of the target domain in general.

\textbf{Pixel Domain Adaptation.}
When there is a domain shift between the two domains, we have
\begin{equation}\label{eq4}
  P_{\mathcal{S}}(S, B, Z, X) \neq P_{\mathcal{T}}(S, B, Z, X)
\end{equation}
According to Bayes' Formula, we can decompose the joint distribution as:
\begin{equation}\label{eq5}
  P_i(S, B, Z, X) = P_i(S, B | Z, X) P_i(Z, X)
\end{equation}
where $i\in\{\mathcal{S}, \mathcal{T}\}$ denotes the different domains. We make the covariate shift assumption for object tracking. $P(S, B | Z, X)$ denotes the conditional distribution, which is the same for source and target domain, i.e.  $P_{\mathcal{S}}(S, B | Z, X) = P_{\mathcal{T}}(S, B | Z, X)$
Therefore, the domain shift is mainly caused by the marginal distribution, namely
\begin{equation}\label{eq7}
  P_{\mathcal{S}}(Z, X)\neq P_{\mathcal{T}}(Z, X)
\end{equation}
In ideal circumstances, the predicted results of SiamRPN++ should be the same regardless of which domain the sequence belongs to. According to the network framework, $P(S, B | Z, X)$ represents RPN for classification and regression. $P(Z, X)$ denotes the feature maps of \emph{template} and \emph{search region} that are the output of the Siamese network. The ultimate aim is to reinforce the Siamese network to extract domain-invariant feature maps, such that,
\begin{equation}\label{eq8}
\begin{split}
P_{\mathcal{S}}(Z, X)= P_{\mathcal{T}}(Z, X)
\end{split}
\end{equation}

\textbf{Semantic Domain Adaptation.}
Pixel domain adaptation solves the domain shift caused by weather or illumination, refers to global region. However the tracking targets of different domain data suffer from appearance change and category variation. The semantic information of the tracking target should also be taken seriously. Similar to pixel domain adaptation, we can decompose the joint distribution into another format by Bayes' Formula, formulated as:
\begin{equation}\label{eq9}
  P(S, B, Z, X) = P(S | B, Z, X) P(B, Z, X)
\end{equation}
We also make the covariate shift assumption for semantic domain adaptation. Assuming that the conditional distribution $P(S | B, Z, X)$ is the same for different domains, i.e,
\begin{equation}\label{eq10}
  P_{\mathcal{S}}(S | B, Z, X) = P_{\mathcal{T}}(S | B, Z, X)
\end{equation}
According to Eq.(\ref{eq9}), we argue that the marginal distribution $P(B, Z, X)$ is the main reason that leads to the domain discrepancy and performance reduction. Therefore, we should constrain the marginal distribution to be the same for the two domains, ideally $P_{\mathcal{S}}(B, Z, X)=P_{\mathcal{T}}(B, Z, X)$. Note that the tracking target annotations are unavailable for the target domain. Therefore, whether source or target domain, we use the predicted bounding boxes from RPN.

\section{The Proposed Method}
\label{sec5}
\begin{figure*}
\begin{center}
\includegraphics[scale=0.56]{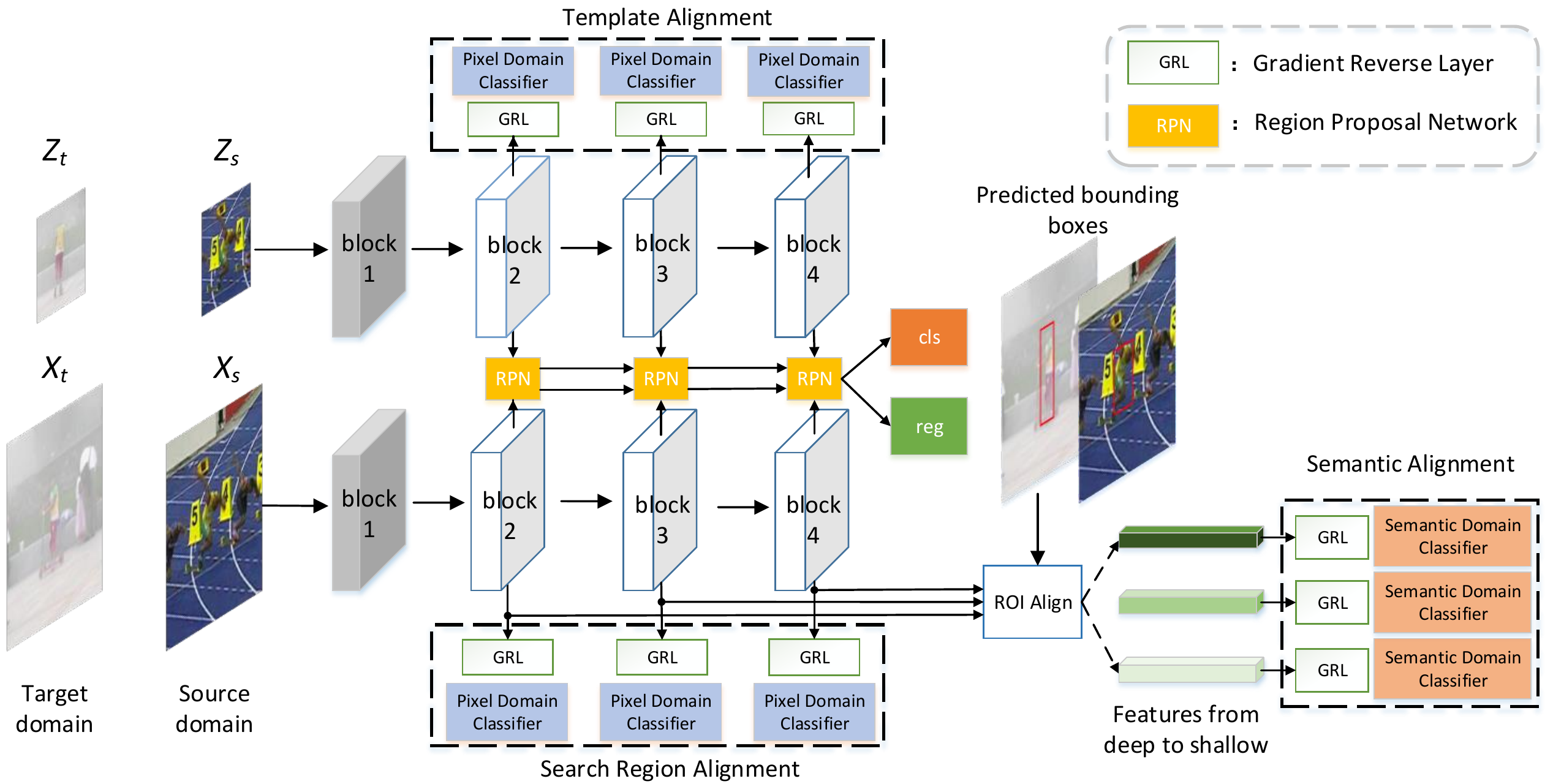}

\end{center}
   \caption{Overview of the proposed framework. The PDA modules including the template alignment and search region alignment are implemented on different \emph{blocks} of Siamese network to make the multi-level feature maps similar across domains.
   Based on the predicted bounding boxes and multi-level feature maps, ROI Align layers can obtain the target feature representations full of sematic information. The SDA modules are adopted to perform the feature alignment on the tracking target between domains.}
   \vspace{-0.3cm}
\label{framework}
\end{figure*}
\subsection{Siamese Network}
The whole framework is illustrated in Fig. \ref{framework}.
Inspired by the $\mathcal{A}$-distance theory and the probabilistic perspective in Section \ref{sec4}, we introduce two domain adaptive modules to solve the domain shift problem, including Pixel Domain Adaptation (PDA) and Semantic Domain Adaptation (SDA).
The modified ResNet-50 \cite{he2016deep} is employed as our backbone network for feature extraction. The stride of original ResNet-50 is 32 pixels. To make it better suitable for tracking, we adjust the effective strides of \emph{block} 3 and 4 from 16 pixels and 32 pixels to 8 pixels by following the same protocol as SiamRPN++ \cite{li2019siamrpn++}. Due to different levels of Siamese network contain low-level and high-level feature maps with diverse feature representations, we intend to leverage multi-level features to take full advantage of semantic information and appearance information.
\subsection{Pixel Domain Adaptation}
As discussed in Section \ref{4.2}, the feature maps, which are extracted by the Siamese network, contain the foreground and background information of the \emph{template} and \emph{search region}. The training images are diverse in image style, illumination and so on. Therefore, to obtain the domain-invariant feature maps, the pixel domain adaptation module includes the \emph{template alignment} and the \emph{search region alignment}, which aims to confuse the feature maps across domains, with minimax optimization between the domain classifiers and the Siamese network.

Given an image pair, i.e. the \emph{template} $z_i$ and the \emph{search region} $x_i$ from the source or target domain, we first extract the multi-level feature maps from \emph{block} 2, 3, 4, which are denoted as $\varphi_k(z)$ and $\varphi_k(x)$, $k\in\{2,3,4\}$. We therefore intend to constrain the feature maps of each level by a domain classifier. The pixel domain classifier is composed of a convolutional layer, a maxpooling layer and a fully connected layer. The FC layer is used for binary classification of each feature pixel. The pixel domain adaptation loss is formulated as:
\begin{equation}\label{eq13}
\small
  \begin{split}
  L_{pda} = -\frac{1}{MN} \sum_{m,n,i}^{M,N} [D_i^{(m,n)} \log p_i^{(m,n)} + \\
  (1-D_i^{(m,n)}) \log (1-p_i^{(m,n)})]
  \end{split}
\end{equation}
where $p_i^{(m,n)}$ stands for the domain classifier's output located at pixel $(m,n)$. $D_i^{(m,n)} \in \{0, 1\}$ denotes the domain label for the source or target domain.
We intend that the domain classifier cannot distinguish whether the pixel belongs to the source domain or not. Following the idea of Eq.(\ref{eq3}), we first minimize the domain classification loss for the domain classifier learning and maximize this loss for the Siamese network learning. The trainable parameters of the domain classifiers are denoted as $w_{pda}$, while $\varphi$ represents the parameters of the Siamese network. The adversarial learning can be written as:
\begin{equation}\label{eq14}
  \max_{\varphi}\min_{w_{pda}}L_{pda}
\end{equation}

To this end, GRL is arranged between the domain classifiers and the Siamese network, as shown in the dashed boxes of Fig. \ref{framework}. In the backpropagation stage, the parameters updating direction of the domain classifier is the same as the direction of reducing the domain classification loss, which is the same as the ordinary training approach. The parameters updating direction of the Siamese network is reversed after GRL, which is exactly the direction of increasing the domain classification loss. After the adversarial training, the feature maps extracted by the Siamese network tend to have similar representations between domains. Therefore, the trackers can perform well on the target domain data.

\subsection{Semantic Domain Adaptation}
Because of the variation of category, angle of view and posture of different domains, the appearance of tracking target will change significantly. Therefore, we propose the Semantic Domain Adaptation (SDA) module to enforce the feature representations of the tracking targets to be domain-invariant in semantic-level.

Along with the outputs of RPN, we can obtain the predicted bounding boxes. ROI Align \cite{he2017mask} takes the predicted bounding boxes and multi-layer feature maps $\varphi_k(x), k \in \{2,3,4\}$ as input. The outputs of ROI Align are the target features with the fixed size of $5 \times 5 \times C$. To align the feature representations of the local region between the two domains, we adopt the adversarial training strategy that is similar to the pixel domain adaptation. GRL is implemented between the domain classifiers and the ROI Align. Considering to harmonize the semantic features from the aspect of the whole target, we utilise two fully connected layers to constitute the semantic domain classifiers. The cross-entropy loss is adopted, i.e.
\begin{equation}\label{eq15}
  \begin{split}
  L_{sda} = -\sum_{i}{[D^k_i \log s^k_i + } {(1-D^k_i) \log (1-s^k_i)]}
  \end{split}
\end{equation}
where $s^k_i$ denotes the outputs of the semantic domain classifier. We also train the loss function of the SDA module in an adversarial manner,
\begin{equation}\label{eq16}
  \max_{\varphi}\min_{w_{sda}}L_{sda}
\end{equation}
where $\varphi$ is the parameters of the Siamese network and $w_{sda}$ denotes the parameters of the semantic domain classifier. Whether the tracking target is from the source or target domain, the domain-invariant feature of the target can obtain a high response in the score map.
\begin{figure*}
    \begin{center}
    \includegraphics[scale=0.47]{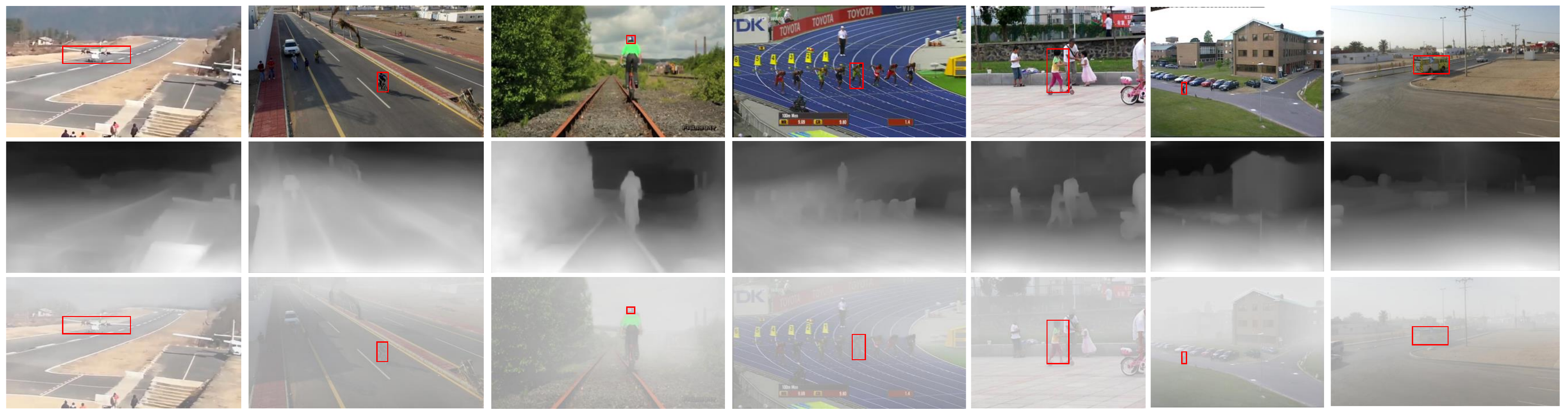}
    \end{center}
       \caption{Illustration of the synthetic foggy images. Original normal images, single-view depth prediction and synthetic foggy images are shown from the first row to the last row. Bounding boxes in red denote the tracking targets.}
       \vspace{-0.3cm}
    \label{foggy_image}
\end{figure*}
\subsection{Training Loss}
 In the training phase, the whole framework is composed of the baseline tracker and domain adaptive modules. Therefore, the final loss of the proposed framework consists of the tracking loss $L_{t}$ and the domain adaptation loss,
\begin{equation}\label{eq17}
  L = L_{t} + \lambda_{da} \sum_{k=2}^{4}(L_{pda}^k+L_{sda}^k)
\end{equation}
where $L_{t}$ consists of the cross-entropy based binary classification loss for foreground-background prediction and the smooth $L$1 based regression loss for object location. $\lambda_{da}$ is a trade-off parameter. $k$ denotes the index of multi-level feature maps.
\section{Foggy Image Generation}
 To validate the domain shift problem, we intend to take the normal sequences as the source domain and the foggy sequences as the target domain. However, there is no public foggy tracking dataset up to now and bounding boxes annotation is also expensive and time-consuming. Using synthetic data to train CNNs becomes increasingly popular. For example, synthetic foggy images are widely utilized in data-driven dehazing methods and domain adaptive detection. Inspired by \cite{chen2018domain}, we propose to utilize existing tracking datasets to synthesize foggy sequences.
 The images in the existing tracking datasets have three channels (RGB). In order to make the synthetic foggy datasets more realistic. Following pioneer works \cite{li2018benchmarking}, the method of synthesizing foggy images can be divided two steps: (1) Single-view depth prediction algorithms are used to generate depth maps; (2) According to the optical model of foggy weather, we utilize the RGB images and their corresponding depth maps to synthesize the foggy images. The synthetic foggy images are shown in Fig. \ref{foggy_image}.

 For the single-view depth prediction model, we select the commonly used MegaDepth \cite{li2018megadepth}. Benefit from large amount of Internet SfM+MVS data, MegaDepth exhibits high accuracy and great generalization performance. The predicted depth maps are shown in the second row of Fig. \ref{foggy_image}. Afterwards, RGB-Depth images are used to generate foggy images.
 According to Mie scattering theory \cite{mccartney1976optics} and HazeRD \cite{zhang2017hazerd}, the airlight and the attenuation are the two main factors of modeling the image information under haze.
Due to the presence of dust, mist or fumes in the air, when the reflected light of the object travels in the air to reach the camera, the intensity of the reflected light will decay exponentially with the increase of distance. The atmospheric scattering model has been the classical description for the foggy image generation, which can be formulated as:
\begin{equation}\label{eq18}
  I_\lambda = t_\lambda E_\lambda + (1-t_{\lambda})A_\lambda
\end{equation}
where $\lambda\ $denotes the wavelength, $E_\lambda$ is the spectral irradiance, $A_\lambda$ represents the airlight, and $t_\lambda$ is the transmission which can be expressed as:
\begin{equation}\label{eq19}
  t_\lambda = e^{-d\beta_\lambda}
\end{equation}
where $\beta_\lambda$ is the scattering coefficient for the haze particles and $d$ denotes the distance between the object and the camera. Foggy VOT2018 \cite{Kristan2018a}, Foggy OTB100 \cite{7001050}, Foggy UAV123 \cite{mueller2016benchmark} and Foggy GOT-10k \cite{huang2019got} benchmarks are generated in this way. It is worth mentioning that we adopt different $\beta_\lambda$ for the training and evaluation datasets, in order to prevent the problem of overfitting.

\section{Experiments}

\subsection{Implementation Details}
\textbf{Data Preprocessing. }
We perform the same data preparation as SiamRPN++. For the source training dataset, i.e. LaSOT \cite{fan2019lasot}, annotations can be used to get the cropped images. However, for the target training dataset, i.e. Foggy GOT-10k \cite{huang2019got} and LSOTB-TIR \cite{liu2020lsotb}, the annotations are not available. In practice, we should first run the existing SiamRPN++ on the target domain datasets to obtain pseudo labels, which can be used to crop images.

\textbf{Training. } We use the modified ResNet-50 \cite{he2016deep} pre-trained on ImageNet \cite{krizhevsky2012imagenet}  as our backbone. In the training stage,  we train the model for 19 epochs with mini-batches of size 16. For the first 10 epochs, we freeze the parameters of backbone and only fine-tune the parameters of three RPN heads \cite{ren2015faster}. For the last 9 epochs, the parameters of \emph{block 2}, 3 and 4 are unfrozen. The base learning rate is $5 \times 10^{-3}$ and we use a warm-up learning rate of $10^{-3}$ for the first 5 epochs. For the last 14 epochs,  the learning rate is decreased exponentially at each epoch from $5 \times 10^{-3}$ to $5 \times 10^{-4}$. We train the network by stochastic gradient descent (SGD) with a weight decay of $10^{-4}$ and a momentum of $0.9$. For domain adaptive modules, their initial learning rate is set to $10^{-3}$ and decreased as the learning rate of RPN. The experiments are implemented by PyTorch on PC with an Intel i9-7900X and four NVIDIA TITAN Xp GPUs.

\begin{table}[t]
    \centering
    \footnotesize
    \setlength{\tabcolsep}{1.1mm}{
    \begin{tabular}{c|ccc|ccc|c}
        \hline
        \hline
        \multicolumn{1}{c}{ \multirow{2}*{Epoch} }& \multicolumn{3}{|c|}{SiamRPN++} & \multicolumn{3}{c|}{DASiamRPN++} & \multicolumn{1}{c}{ \multirow{2}*{$\Delta_{EAO}$($\%$)} }\\
        \cline{2-7}
        \multicolumn{1}{c|}{}&A$\uparrow$&R$\downarrow$&EAO$\uparrow$&A$\uparrow$&R$\downarrow$&EAO$\uparrow$\\
        \hline
        \hline
        Epoch 11     &0.534&0.576&0.211    &0.544&0.557&\textbf{0.218}&0.7\\
        Epoch 12     &0.539&0.759&0.164    &0.547&0.660&\textbf{0.187}&2.3\\
        Epoch 13     &0.558&0.782&0.162    &0.550&0.548&\textbf{0.206}&4.4\\
        Epoch 14     &0.548&0.759&0.160    &0.547&0.510&\textbf{0.231}&7.1\\
        Epoch 15     &0.553&0.763&0.166    &0.541&0.642&\textbf{0.186}&2.0\\
        Epoch 16     &0.537&0.833&0.150    &0.539&0.585&\textbf{0.208}&5.8\\
        Epoch 17     &0.555&0.810&0.157    &0.542&0.609&\textbf{0.203}&4.6\\
        Epoch 18     &0.544&0.871&0.145    &0.523&0.604&\textbf{0.193}&4.8\\
        Epoch 19     &0.540&0.698&0.176    &0.536&0.618&\textbf{0.193}&1.7\\
        \hline
        \hline
    \end{tabular}}
    \vspace{0.2cm}
    \caption{Quantitative results on Foggy VOT2018 \cite{Kristan2018a}}
    \vspace{-0.1cm}
    \label{Foggy_vot_table}
\end{table}

\begin{table}
    \centering
    \footnotesize

    \setlength{\tabcolsep}{2.1mm}{
    \begin{tabular}{c|cc|cc|cc}
        \hline
        \hline
        \multicolumn{1}{c}{ \multirow{2}*{Epoch} }&
        \multicolumn{2}{|c|}{SiamRPN++} &
        \multicolumn{2}{c|}{DASiamRPN++} &
        \multicolumn{1}{c}{\multirow{2}*{$\Delta_p$}} &
        \multicolumn{1}{c}{\multirow{2}*{$\Delta_s$ ($\%$)}}\\
        \cline{2-5}
        \multicolumn{1}{c|}{}&P$\uparrow$&S$\uparrow$&P$\uparrow$&S$\uparrow$\\
        \hline
        \hline
        Epoch 11     &0.737&\textbf{0.533}    &0.726&\textbf{0.533}&-1.1&0\\
        Epoch 12     &0.736&\textbf{0.533}    &0.715&0.516&-2.2&-1.7\\
        Epoch 13     &0.725&\textbf{0.527}    &0.725&0.525&0&-0.2\\
        Epoch 14     &0.699&0.508    &0.740&\textbf{0.540}&4.1&3.2\\
        Epoch 15     &0.712&0.522    &0.736&\textbf{0.528}&2.4&0.6\\
        Epoch 16     &0.734&0.531    &0.740&\textbf{0.541}&0.6&1\\
        Epoch 17     &0.714&0.516    &0.740&\textbf{0.544}&3.6&2.8\\
        Epoch 18     &0.694&0.507    &0.711&\textbf{0.525}&1.7&1.8\\
        Epoch 19     &0.701&0.513    &0.728&\textbf{0.539}&2.7&2.6\\
        \hline
        \hline
    \end{tabular}}
    \vspace{0.2cm}
    \caption{Quantitative results on Foggy OTB100 \cite{7001050}.}
    \vspace{-0.1cm}
    \label{Foggy_otb_table}
\end{table}

\begin{table}
    \centering
    \footnotesize

    \setlength{\tabcolsep}{2.1mm}{
    \begin{tabular}{c|cc|cc|cc}
        \hline
        \hline
        \multicolumn{1}{c}{ \multirow{2}*{Epoch} }&
        \multicolumn{2}{|c|}{SiamRPN++} &
        \multicolumn{2}{c|}{DASiamRPN++} &
        \multicolumn{1}{c}{ \multirow{2}*{$\Delta_p$}}&
        \multicolumn{1}{c}{ \multirow{2}*{$\Delta_s$ ($\%$)}}\\
        \cline{2-5}
        \multicolumn{1}{c|}{}&P$\uparrow$&S$\uparrow$&P$\uparrow$&S$\uparrow$\\
        \hline
        \hline
        Epoch 11     &0.671&\textbf{0.487}    &0.670&0.476&-0.1&-1.1\\
        Epoch 12     &0.664&0.472    &0.670&\textbf{0.477}&0.6&0.5\\
        Epoch 13     &0.641&0.457    &0.685&\textbf{0.491}&4.4&3.4\\
        Epoch 14     &0.675&0.473    &0.687&\textbf{0.487}&1.2&1.4\\
        Epoch 15     &0.659&0.467    &0.685&\textbf{0.482}&2.6&1.5\\
        Epoch 16     &0.650&0.458    &0.702&\textbf{0.498}&4.8&4\\
        Epoch 17     &0.645&0.454    &0.677&\textbf{0.484}&3.2&3\\
        Epoch 18     &0.644&0.453    &0.661&\textbf{0.471}&1.7&1.8\\
        Epoch 19     &0.661&0.467    &0.664&\textbf{0.475}&0.3&0.8\\
        \hline
        \hline
    \end{tabular}}
    \vspace{0.2cm}
    \caption{Quantitative results on Foggy UAV123 \cite{mueller2016benchmark}.}
    \vspace{-0.1cm}
    \label{Foggy_uav_table}
\end{table}

\subsection{Cross-domain Tracking from Normal to Foggy}
\textbf{Foggy VOT2018 \cite{Kristan2018a}:} VOT dataset consists of 60 challenging sequences and the VOT challenges update some sequences annually.
The performance is evaluated in terms of Accuracy (A) and Robustness (R) and EAO, which respectively denote the average overlap over successfully tracked frames, failure rate and Expected Average Overlap.

In order to better reflect the performance improvements of our method, we select the trained models of SiamRPN++ and DASiamRPN++ from \emph{Epoch} 11 to 19, which are evaluated on Foggy VOT2018 without fine-tuning any hyper-parameters. We first compare the best results of DASiamRPN++ and SiamRPN++. As presented in Table \ref{Foggy_vot_table}, 
our DASiamRPN++ can achieve the best EAO score of $0.231$, which outperforms the best EAO of SiamRPN++ $0.211$ with a relative gain of $9.5\%$. For the quantitative results of each epoch in Table \ref{Foggy_vot_table}, our method can surpass SiamRPN++.

\textbf{Foggy OTB100 \cite{7001050}:} 
Foggy OTB100 contains 100 sequences that are collected from common tracking sequences. The evaluation is based on two indicators: precision (P) and success (S) scores.
The results of each epoch are presented in Table \ref{Foggy_otb_table}. The best success score of SiamRPN++ is $0.533$. Compared with SiamRPN++, our DASiamRPN++ achieves $+2.1\%$ performance gain using PDA and SDA modules. For each epoch, most models of our method surpass SiamRPN++, which demonstrates the effectiveness of our method.

\begin{table}
    \centering
    \footnotesize
    \setlength{\tabcolsep}{2.1mm}{
    \begin{tabular}{c|cc|cc|cc}
        \hline
        \hline
        \multicolumn{1}{c}{ \multirow{2}*{Epoch} }&
        \multicolumn{2}{|c|}{SiamRPN++} &
        \multicolumn{2}{c|}{DASiamRPN++} &
        \multicolumn{1}{c}{ \multirow{2}*{$\Delta_p$}}&
        \multicolumn{1}{c}{ \multirow{2}*{$\Delta_s$ ($\%$)}}\\
        \cline{2-5}
        \multicolumn{1}{c|}{}&P$\uparrow$&S$\uparrow$&P$\uparrow$&S$\uparrow$\\
        \hline
        \hline
        Epoch 11     &0.620&0.511    &0.636&\textbf{0.528}&1.6&1.7\\
        Epoch 12     &0.638&0.527    &0.644&\textbf{0.532}&0.6&0.5\\
        Epoch 13     &0.643&0.526    &0.661&\textbf{0.541}&1.8&1.5\\
        Epoch 14     &0.655&0.541    &0.665&\textbf{0.547}&1.0&0.6\\
        Epoch 15     &0.645&0.537    &0.658&\textbf{0.543}&1.3&0.6\\
        Epoch 16     &0.631&0.518    &0.655&\textbf{0.535}&2.4&1.7\\
        Epoch 17     &0.647&\textbf{0.537}    &0.652&0.535&0.5&-0.2\\
        Epoch 18     &0.647&\textbf{0.534}    &0.642&0.528&-0.5&-0.6\\
        Epoch 19     &0.657&\textbf{0.543}    &0.661&0.542&0.4&-0.1\\
        \hline
        \hline
    \end{tabular}}
    \vspace{0.2cm}
    \caption{Quantitative results on LSOTB-TIR test subset \cite{liu2020lsotb}.}
    \vspace{-0.3cm}
    \label{foggy_lstir_res}
\end{table}

\textbf{Foggy UAV123 \cite{mueller2016benchmark}:} UAV123 consists of 123 aerial video sequences,
which take precision (P) and success (S) scores as indicators for performance comparison.
All of the results are shown in Table \ref{Foggy_uav_table}. The best success score of our method is $0.498$. We can achieve a relative gain of $2.3\%$ that compared with the best success score of SiamRPN++. For the comparison of each epoch models, our approach can significantly improve both the precision and success scores.


\begin{figure*}
  \center
  \includegraphics[scale=0.52]{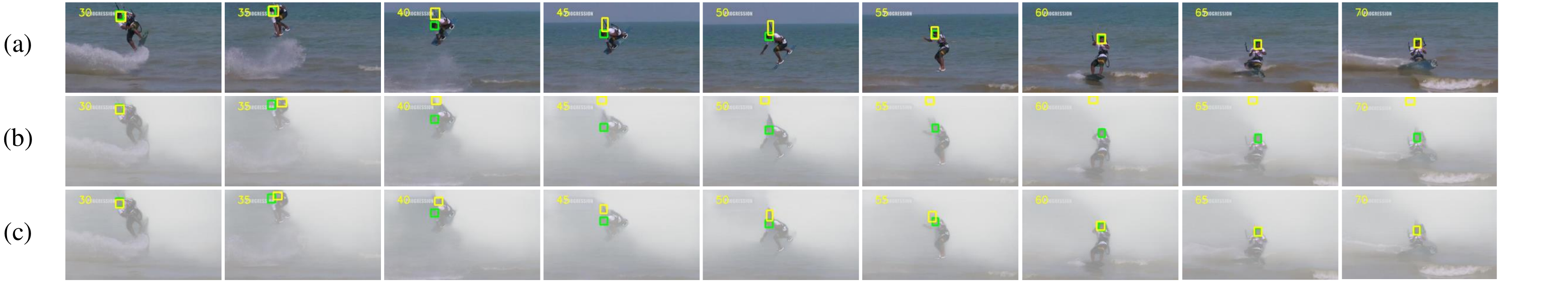}
  \caption{Visualization of the tracking results. (a) denotes the tracking results of SiamRPN++ on the normal sequence. (b) denotes the tracking results of SiamRPN++ on the foggy sequence. (c) denotes the tracking results of DASiamRPN++ on the foggy sequence. Domain shift leads to a tracking failure according to (a) and (b). (b) and (c) show that our method has better transferability cross domains.}

  \label{sequence}
\end{figure*}

\begin{table*}
  \footnotesize
  \centering
  \setlength{\tabcolsep}{1.9mm}{
  \begin{tabular}{c|c|c|c|c|c|c|c|c|c|c|c}
    \hline
    \hline
    Tracker   &  PDA & SDA & Epoch 11 & Epoch 12 & Epoch 13 & Epoch 14 & Epoch 15 & Epoch 16 & Epoch 17 & Epoch 18&Epoch 19 \\
    \hline
    SiamRPN++   &              &                & 0.211 & 0.164 & 0.162 & 0.160 & 0.166 & 0.150 & 0.157 & 0.145 & 0.176 \\
    Ours        & $\checkmark$ &                & 0.186 & 0.179 & 0.187 & 0.186 & 0.171 & 0.187 & 0.164 & 0.178 & 0.183 \\
    Ours        &              &  $\checkmark$  & 0.195 & 0.172 & 0.182 & 0.170 & 0.162 & 0.176 & 0.170 & 0.178 & 0.161 \\
    Ours        & $\checkmark$ &  $\checkmark$  & 0.218 & 0.187 & 0.206 & 0.231 & 0.186 & 0.208 & 0.203 & 0.193 & 0.193 \\
    \hline
    \hline
  \end{tabular}}
  \vspace{0.1cm}
  \caption{Ablation study on Foggy VOT2018 \cite{Kristan2018a} in terms of EAO. }
  \vspace{-0.3cm}
  \label{ablation study}
\end{table*}
\subsection{Cross-domain Tracking from RGB to TIR}
\textbf{LSOTB-TIR\cite{liu2020lsotb}:} LSOTB-TIR consists of 1,400 TIR sequences which are divided into training and test subset. The test subset contains 120 videos. LaSOT is selected as the source domain dataset, while we choose the LSOTB-TIR training subset as the target domain dataset. The performance indicators are the same as OTB100. We evaluate SiamRPN++ and DASiamRPN++ on the LSOTB-TIR test subset. The results of each epoch are shown in Table \ref{foggy_lstir_res}. Our method has improvement compared with the baseline.

\subsection{Ablation Study and Discussion}

\begin{figure}[t]
  \centering
  \subfigure[\emph{road} w/o DA]{
    \includegraphics[width=1.7cm]{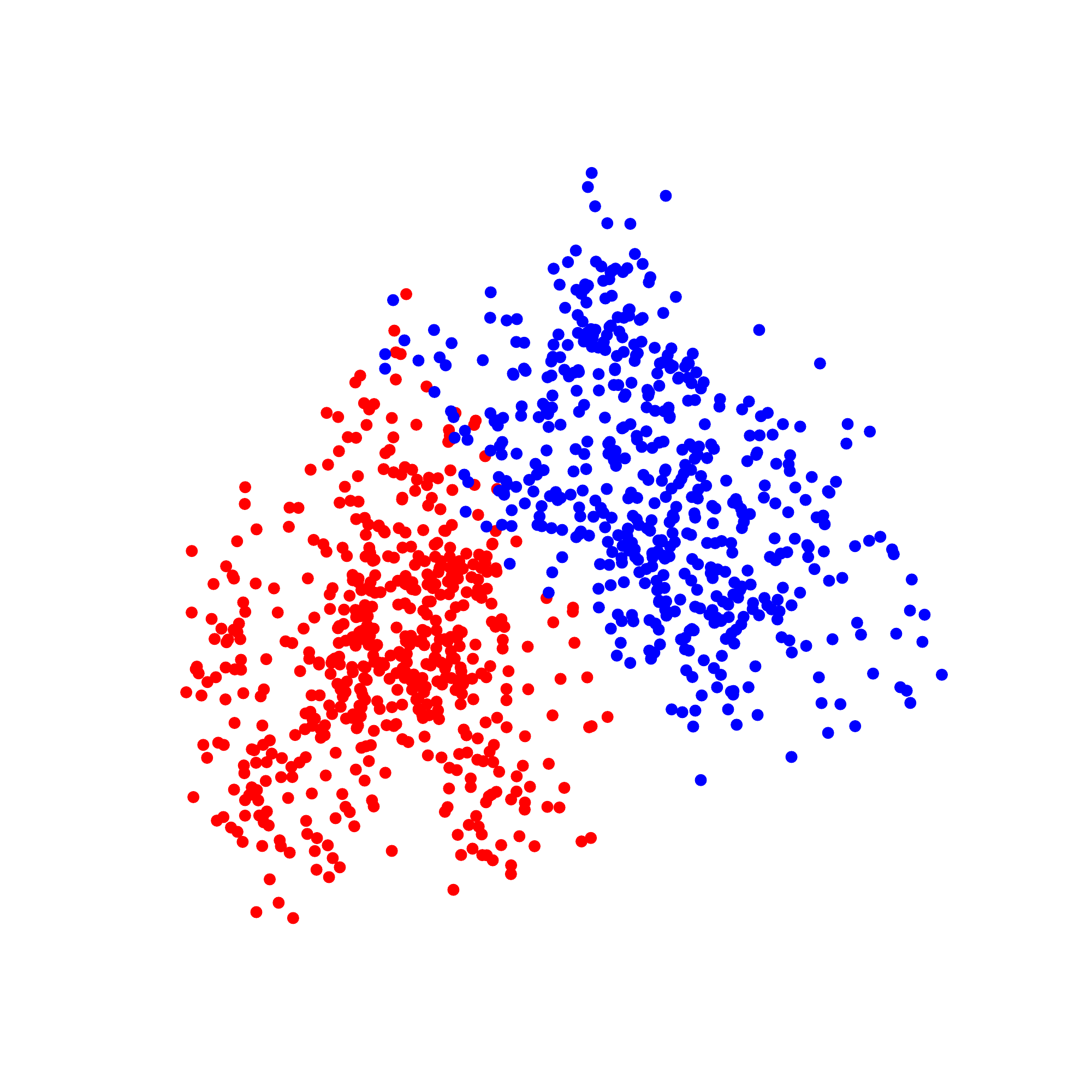}}
  \hspace{2mm}
  \subfigure[\emph{car}1 w/o DA]{
    \includegraphics[width=1.7cm]{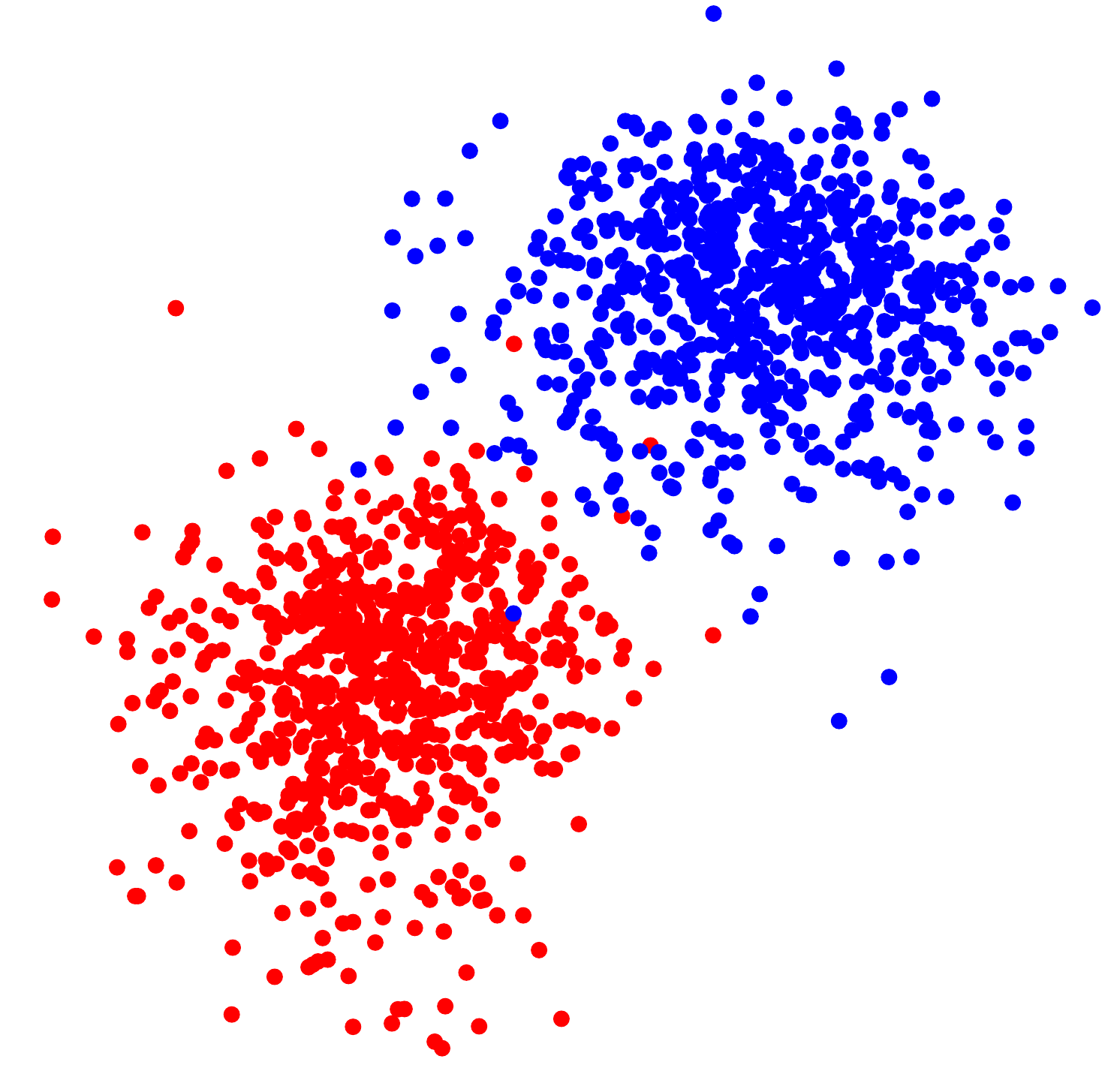}}
  \hspace{2mm}
  \subfigure[\emph{girl} w/o DA]{
    \includegraphics[width=1.7cm]{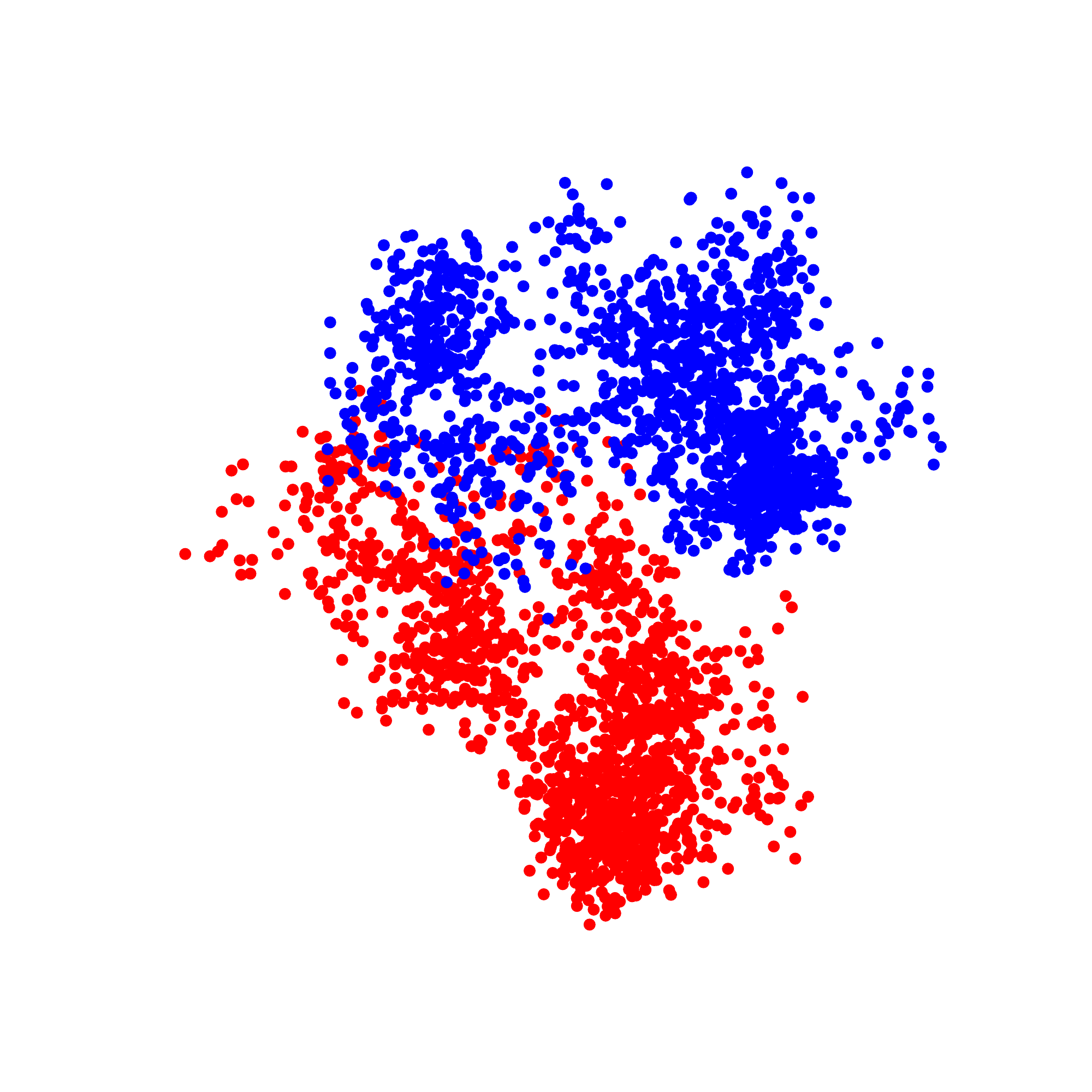}}
  \hspace{2mm}
  \subfigure[\emph{bolt}1 w/o DA]{
    \includegraphics[width=1.7cm]{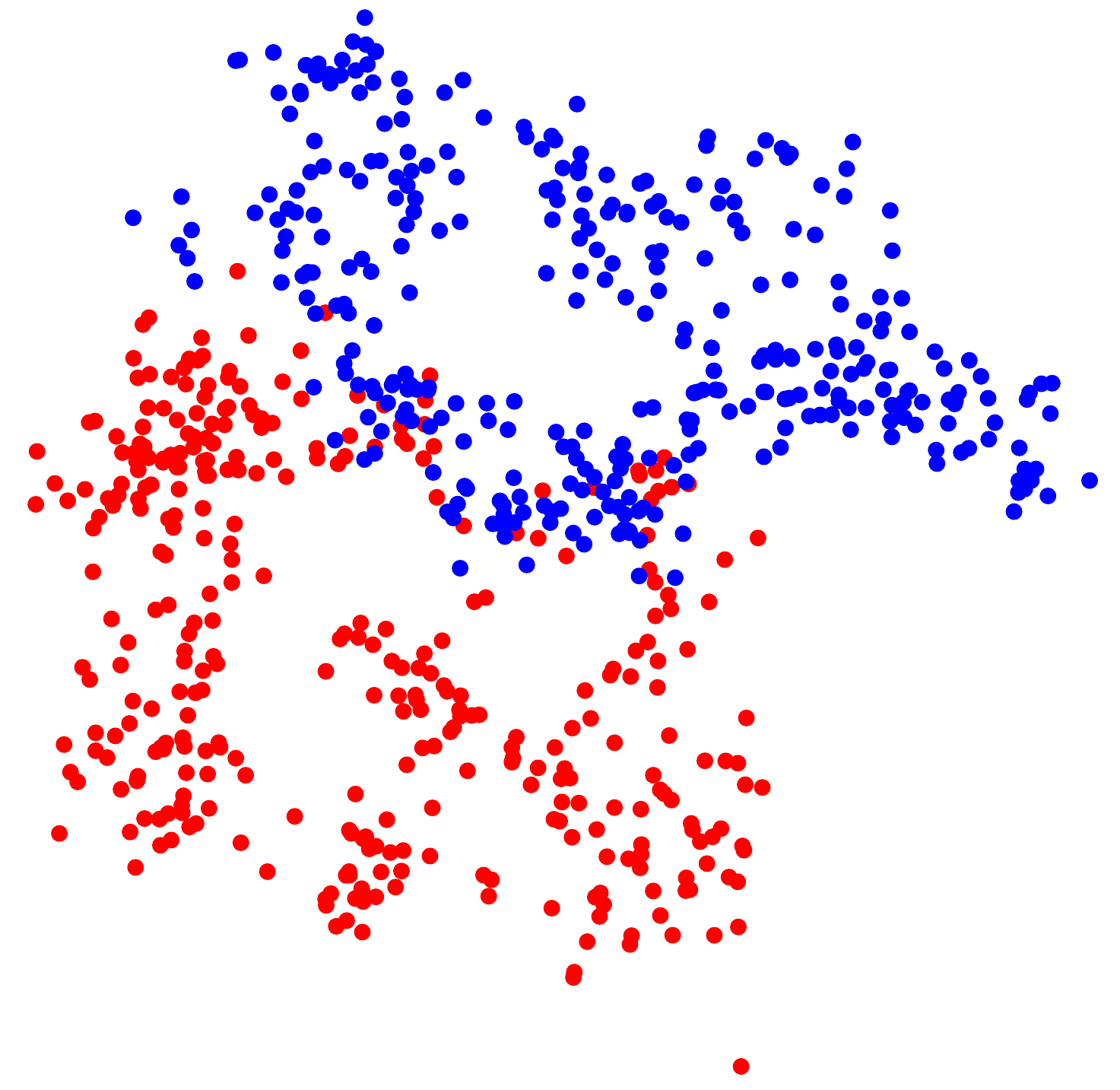}}
  \subfigure[\emph{road} with DA]{
    \includegraphics[width=1.7cm]{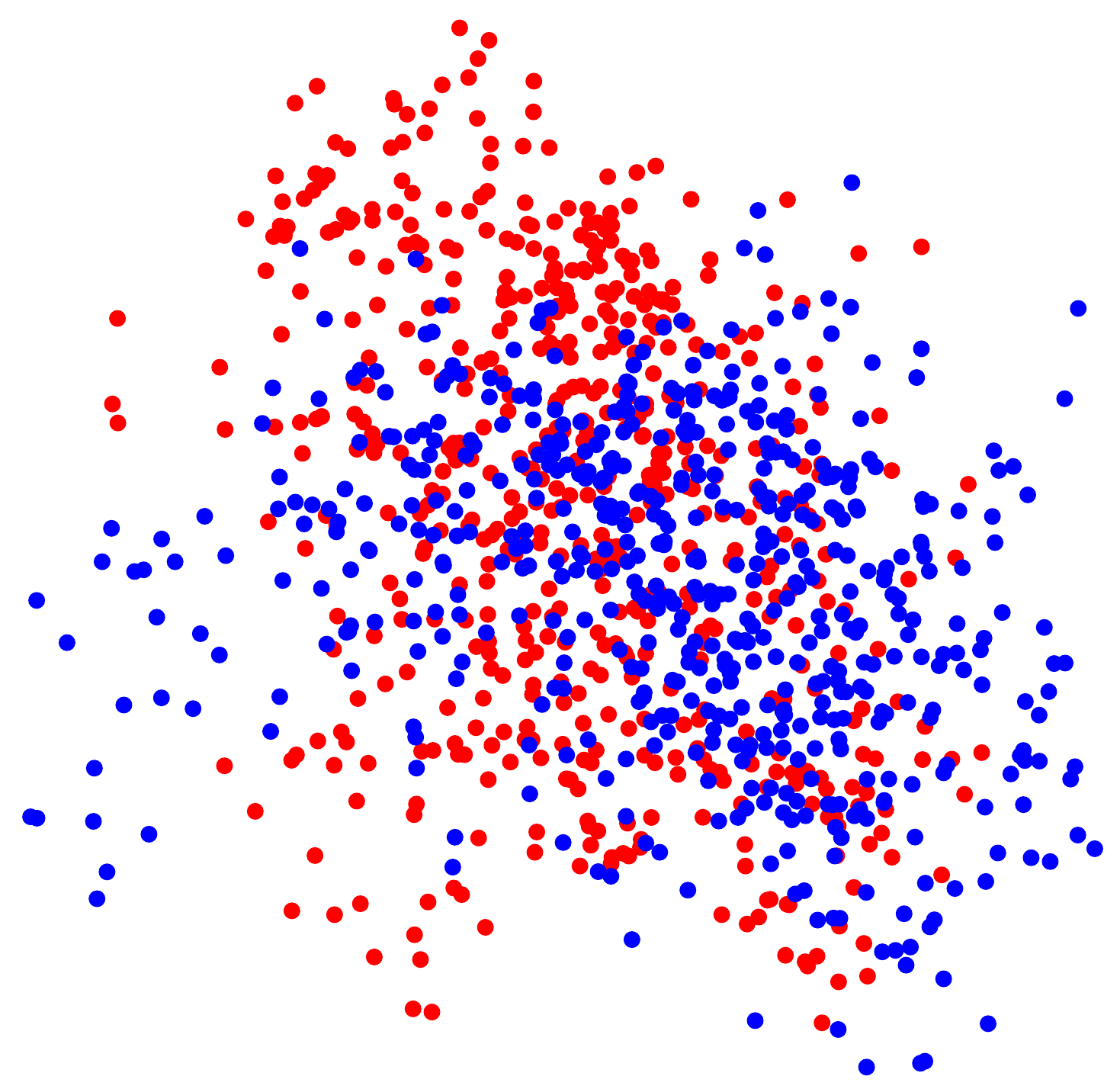}}
  \hspace{2mm}
  \subfigure[\emph{car}1 with DA]{
    \includegraphics[width=1.7cm]{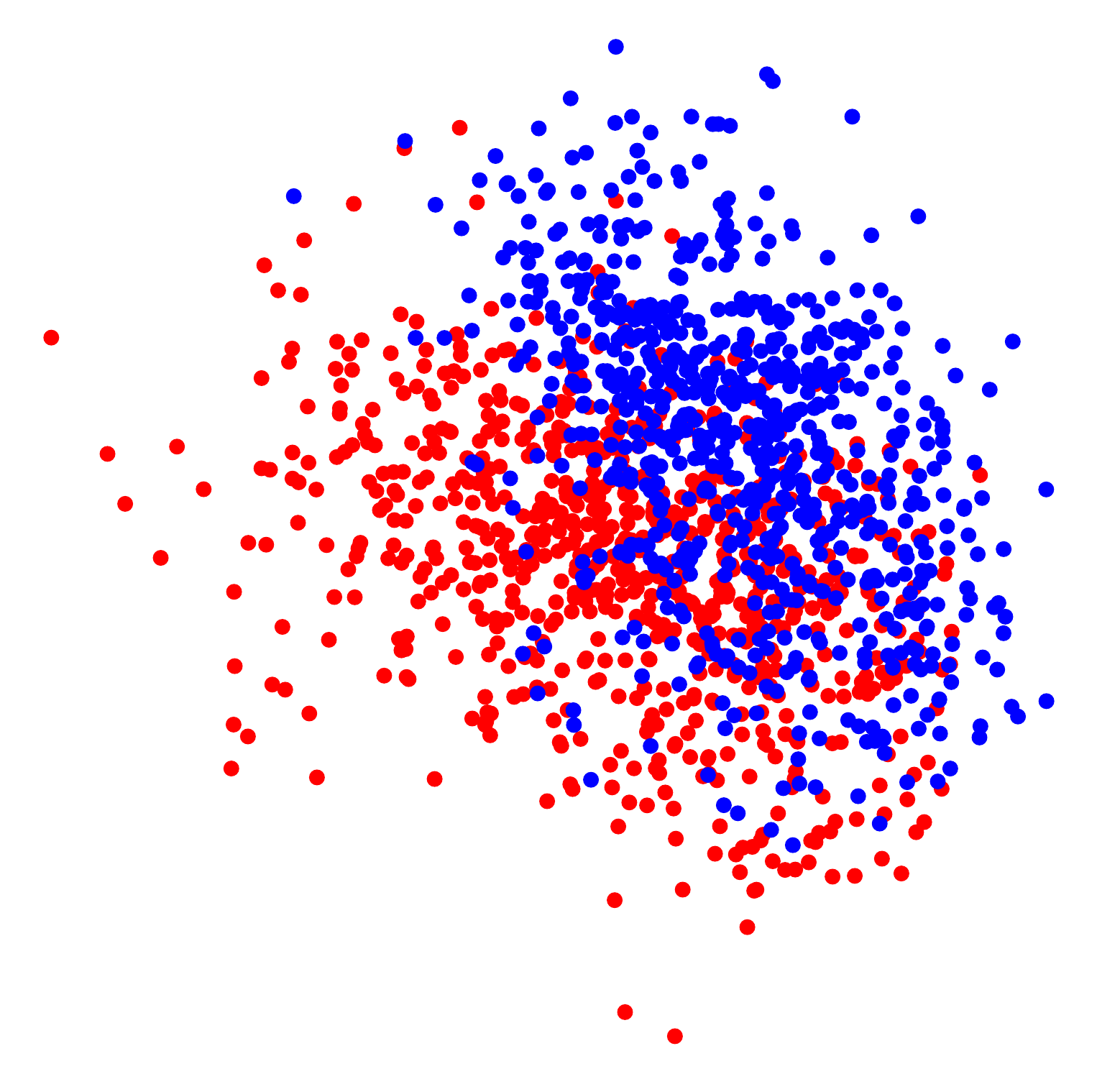}}
  \hspace{2mm}
  \subfigure[\emph{girl} with DA]{
    \includegraphics[width=1.7cm]{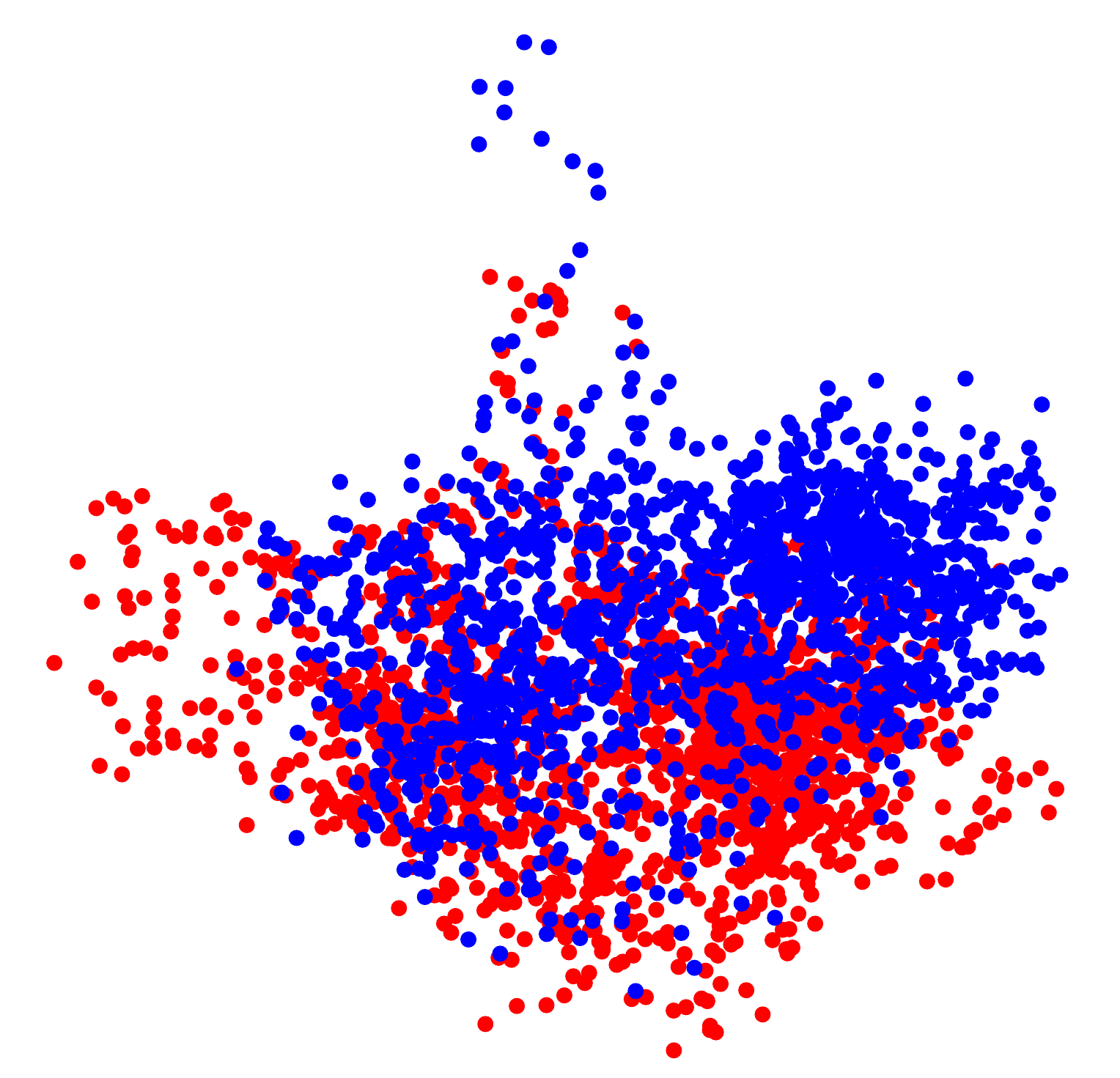}}
  \hspace{2mm}
  \subfigure[\emph{bolt}1 with DA]{
    \includegraphics[width=1.7cm]{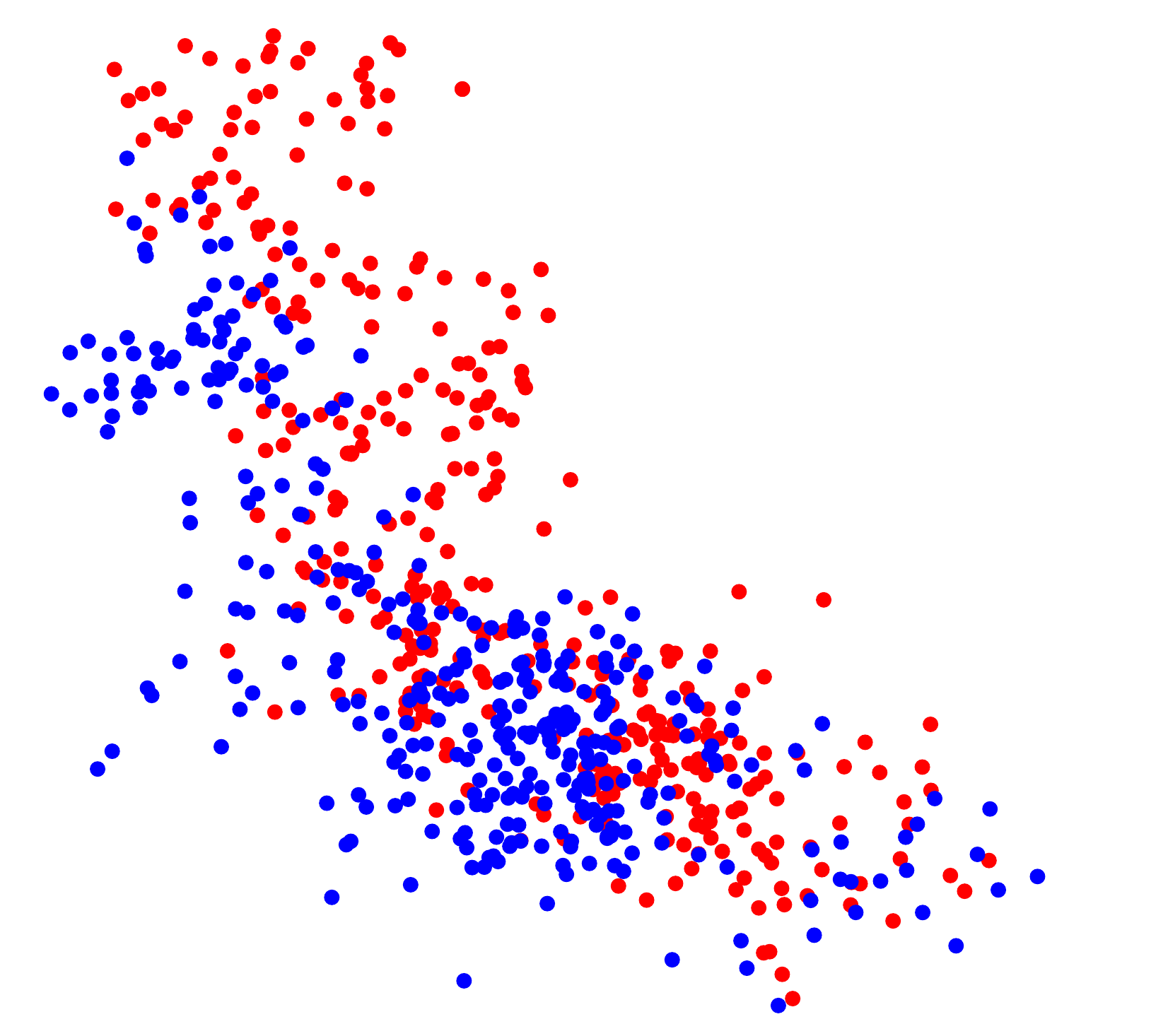}}
  \caption{Feature visualization of the VOT2018 sequences (\emph{road}, \emph{car}1, \emph{girl}, \emph{bolt}1). Red and blue points correspond to the feature maps of the normal and foggy images, respectively. The feature maps extracted by the Siamese network without DA are visualized in the first row. The second row denotes the feature maps extracted by the network with DA.}
  \label{feat_vis}
  \vspace{-0.4cm}
\end{figure}

\textbf{Effectiveness of PDA.}
In Table \ref{ablation study}, we evaluate the models that combine SiamRPN++ and the PDA modules. Obviously, SiamRPN++ with the PDA module can get a better performance.
The PDA module exerts the domain alignment on the feature maps of the \emph{template} and \emph{search region}. The Siamese network is capable of  extracting domain-invariant features to make our tracker more suitable for sophisticated application environments.
For most epoch models, DASiamRPN++ with only PDA module can also surpass the original SiamRPN++.

\textbf{Effectiveness of SDA.}
Similar to aforementioned PDA, our DASiamRPN++ with only SDA are evaluated on Foggy VOT2018 to show the efficiency of the semantic domain adaptation module. In Table \ref{ablation study}, the combination of SiamRPN++ and SDA can achieve better results than SiamRPN++ in different epochs. The proposed SDA modules are helpful to the final domain adaptation.

\textbf{Visualization.}
The t-SNE \cite{maaten2008visualizing} is a frequently-used method for feature dimensionality reduction. It is very suitable for dimensionality reduction of high-dimensional data to 2 or 3 dimensions for easy visualization. The extracted feature maps of \emph{block} 3 are visualized in Fig. \ref{feat_vis}.
By the constraint of domain adaptive modules, the feature maps extracted by DASiamRPN++ are confused. Further, the visualization of the tracking results are shown in Fig. \ref{sequence}.

\begin{figure}
\center
\includegraphics[scale=0.3]{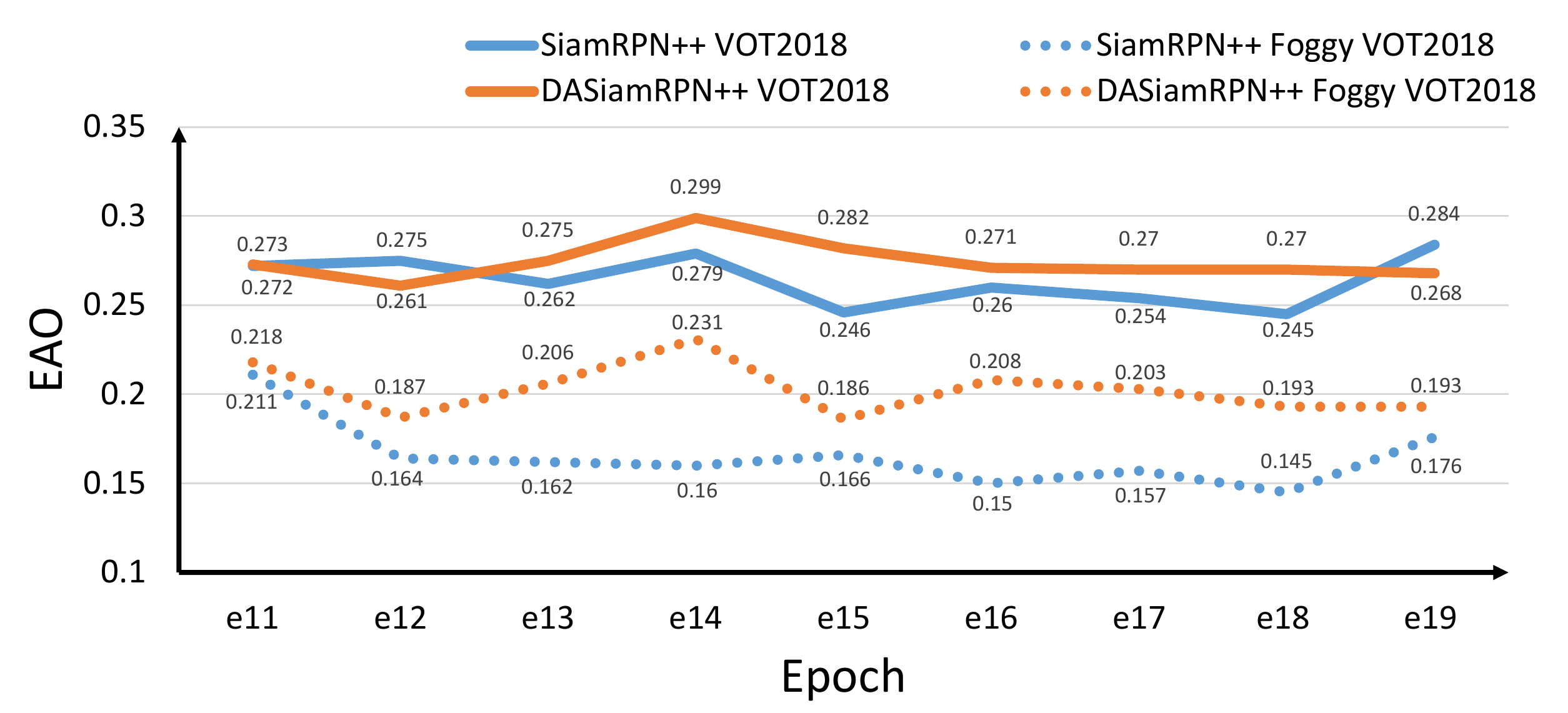}
\caption{Evaluation on Foggy VOT2018 and VOT2018. ``SiamRPN++ VOT2018" denotes that the SiamRPN++ model is evaluated on VOT2018. The others are the same expression.}
\vspace{-0.3cm}
\label{improve}
\end{figure}

\textbf{Evaluation on Normal Sequences. }
 We expect that DASiamRPN++ does not cause any performance drop on the original VOT2018. As is shown in Fig. \ref{improve}, the line in orange denotes the evaluation results of DASiamRPN++, which demonstrates that our method works well on both normal sequences and foggy sequences.

\section{Conclusions}
In this paper, we introduce the problem of domain shift across different datasets into the visual tracking community. To demonstrate that the domain distribution discrepancy will lead to a performance reduction, we design the confirmatory experiments including synthetic foggy sequences and TIR sequences.
Then, in order to solve the domain shift problem, the original normal sequences with full annotations are set as the source domain, while the foggy sequences are set as the target domain which is completely unlabeled. With the $\mathcal{A}$-distance theory and a probabilistic perspective for object tracking, we introduce two domain adaptive modules, namely the pixel domain adaptation module and the semantic domain adaptation module. The two domain adaptive modules are designed to minimize the domain gap across datasets by minimax based adversarial training with GRL. Extensive experiments show that our method can bring significant performance improvements over SiamRPN++, and the better adaptability and transferability for cross-domain tracking are thus demonstrated.

{\small
\bibliographystyle{ieee_fullname}
\bibliography{egbib}

\begin{thebibliography}{10}\itemsep=-1pt

\bibitem{ben2006analysis}
Shai Ben-David, John Blitzer, Koby Crammer, and Fernando Pereira.
\newblock Analysis of representations for domain adaptation.
\newblock {\em Advances in neural information processing systems}, 19:137--144,
  2006.

\bibitem{bertinetto2016fully}
Luca Bertinetto, Jack Valmadre, Joao~F Henriques, Andrea Vedaldi, and Philip~HS
  Torr.
\newblock Fully-convolutional siamese networks for object tracking.
\newblock In {\em European conference on computer vision}, pages 850--865.
  Springer, 2016.

\bibitem{bhat2019learning}
Goutam Bhat, Martin Danelljan, Luc~Van Gool, and Radu Timofte.
\newblock Learning discriminative model prediction for tracking.
\newblock In {\em Proceedings of the IEEE International Conference on Computer
  Vision}, pages 6182--6191, 2019.

\bibitem{bousmalis2017unsupervised}
Konstantinos Bousmalis, Nathan Silberman, David Dohan, Dumitru Erhan, and Dilip
  Krishnan.
\newblock Unsupervised pixel-level domain adaptation with generative
  adversarial networks.
\newblock In {\em Proceedings of the IEEE conference on computer vision and
  pattern recognition}, pages 3722--3731, 2017.

\bibitem{chen2018domain}
Yuhua Chen, Wen Li, Christos Sakaridis, Dengxin Dai, and Luc Van~Gool.
\newblock Domain adaptive faster r-cnn for object detection in the wild.
\newblock In {\em Proceedings of the IEEE conference on computer vision and
  pattern recognition}, pages 3339--3348, 2018.

\bibitem{chen2020siamese}
Zedu Chen, Bineng Zhong, Guorong Li, Shengping Zhang, and Rongrong Ji.
\newblock Siamese box adaptive network for visual tracking.
\newblock In {\em Proceedings of the IEEE/CVF Conference on Computer Vision and
  Pattern Recognition}, pages 6668--6677, 2020.

\bibitem{duan2012domain}
Lixin Duan, Ivor~W Tsang, and Dong Xu.
\newblock Domain transfer multiple kernel learning.
\newblock {\em IEEE Transactions on Pattern Analysis and Machine Intelligence},
  34(3):465--479, 2012.

\bibitem{fan2019lasot}
Heng Fan, Liting Lin, Fan Yang, Peng Chu, Ge Deng, Sijia Yu, Hexin Bai, Yong
  Xu, Chunyuan Liao, and Haibin Ling.
\newblock Lasot: A high-quality benchmark for large-scale single object
  tracking.
\newblock In {\em Proceedings of the IEEE Conference on Computer Vision and
  Pattern Recognition}, pages 5374--5383, 2019.

\bibitem{fan2019siamese}
Heng Fan and Haibin Ling.
\newblock Siamese cascaded region proposal networks for real-time visual
  tracking.
\newblock In {\em Proceedings of the IEEE Conference on Computer Vision and
  Pattern Recognition}, pages 7952--7961, 2019.

\bibitem{ganin2015unsupervised}
Yaroslav Ganin and Victor Lempitsky.
\newblock Unsupervised domain adaptation by backpropagation.
\newblock In {\em International conference on machine learning}, pages
  1180--1189. PMLR, 2015.

\bibitem{goodfellow2014generative}
Ian Goodfellow, Jean Pouget-Abadie, Mehdi Mirza, Bing Xu, David Warde-Farley,
  Sherjil Ozair, Aaron Courville, and Yoshua Bengio.
\newblock Generative adversarial nets.
\newblock In {\em Advances in neural information processing systems}, pages
  2672--2680, 2014.

\bibitem{guo2020siamcar}
Dongyan Guo, Jun Wang, Ying Cui, Zhenhua Wang, and Shengyong Chen.
\newblock Siamcar: Siamese fully convolutional classification and regression
  for visual tracking.
\newblock In {\em Proceedings of the IEEE/CVF Conference on Computer Vision and
  Pattern Recognition}, pages 6269--6277, 2020.

\bibitem{he2017mask}
Kaiming He, Georgia Gkioxari, Piotr Doll{\'a}r, and Ross Girshick.
\newblock Mask r-cnn.
\newblock In {\em Proceedings of the IEEE international conference on computer
  vision}, pages 2961--2969, 2017.

\bibitem{he2016deep}
Kaiming He, Xiangyu Zhang, Shaoqing Ren, and Jian Sun.
\newblock Deep residual learning for image recognition.
\newblock In {\em Proceedings of the IEEE conference on computer vision and
  pattern recognition}, pages 770--778, 2016.

\bibitem{he2019multi}
Zhenwei He and Lei Zhang.
\newblock Multi-adversarial faster-rcnn for unrestricted object detection.
\newblock In {\em Proceedings of the IEEE International Conference on Computer
  Vision}, pages 6668--6677, 2019.

\bibitem{he2020domain}
Zhenwei He and Lei Zhang.
\newblock Domain adaptive object detection via asymmetric tri-way faster-rcnn.
\newblock {\em arXiv preprint arXiv:2007.01571}, 2020.

\bibitem{hoffman2014lsda}
Judy Hoffman, Sergio Guadarrama, Eric~S Tzeng, Ronghang Hu, Jeff Donahue, Ross
  Girshick, Trevor Darrell, and Kate Saenko.
\newblock Lsda: Large scale detection through adaptation.
\newblock In {\em Advances in Neural Information Processing Systems}, pages
  3536--3544, 2014.

\bibitem{huang2019got}
Lianghua Huang, Xin Zhao, and Kaiqi Huang.
\newblock Got-10k: A large high-diversity benchmark for generic object tracking
  in the wild.
\newblock {\em IEEE Transactions on Pattern Analysis and Machine Intelligence},
  2019.

\bibitem{Kristan2018a}
Matej Kristan, Ales Leonardis, Jiri Matas, Michael Felsberg, Roman Pfugfelder,
  Luka \v{C}ehovin Zajc, Tomas Vojir, Goutam Bhat, Alan Lukezic, Abdelrahman
  Eldesokey, Gustavo Fernandez, and et al.
\newblock The sixth visual object tracking vot2018 challenge results, 2018.

\bibitem{Kristan2019a}
Matej Kristan, Jiri Matas, Ales Leonardis, Michael Felsberg, Roman Pflugfelder,
  Joni-Kristian Kamarainen, Luka \v{C}ehovin Zajc, Ondrej Drbohlav, Alan
  Lukezic, Amanda Berg, Abdelrahman Eldesokey, Jani Kapyla, and Gustavo
  Fernandez.
\newblock The seventh visual object tracking vot2019 challenge results, 2019.

\bibitem{krizhevsky2012imagenet}
Alex Krizhevsky, Ilya Sutskever, and Geoffrey~E Hinton.
\newblock Imagenet classification with deep convolutional neural networks.
\newblock In {\em Advances in neural information processing systems}, pages
  1097--1105, 2012.

\bibitem{li2018benchmarking}
Boyi Li, Wenqi Ren, Dengpan Fu, Dacheng Tao, Dan Feng, Wenjun Zeng, and
  Zhangyang Wang.
\newblock Benchmarking single-image dehazing and beyond.
\newblock {\em IEEE Transactions on Image Processing}, 28(1):492--505, 2018.

\bibitem{li2019siamrpn++}
Bo Li, Wei Wu, Qiang Wang, Fangyi Zhang, Junliang Xing, and Junjie Yan.
\newblock Siamrpn++: Evolution of siamese visual tracking with very deep
  networks.
\newblock In {\em Proceedings of the IEEE Conference on Computer Vision and
  Pattern Recognition}, pages 4282--4291, 2019.

\bibitem{li2018high}
Bo Li, Junjie Yan, Wei Wu, Zheng Zhu, and Xiaolin Hu.
\newblock High performance visual tracking with siamese region proposal
  network.
\newblock In {\em Proceedings of the IEEE Conference on Computer Vision and
  Pattern Recognition}, pages 8971--8980, 2018.

\bibitem{li2017mmd}
Chun-Liang Li, Wei-Cheng Chang, Yu Cheng, Yiming Yang, and Barnab{\'a}s
  P{\'o}czos.
\newblock Mmd gan: Towards deeper understanding of moment matching network.
\newblock In {\em Advances in Neural Information Processing Systems}, pages
  2203--2213, 2017.

\bibitem{li2018megadepth}
Zhengqi Li and Noah Snavely.
\newblock Megadepth: Learning single-view depth prediction from internet
  photos.
\newblock In {\em Proceedings of the IEEE Conference on Computer Vision and
  Pattern Recognition}, pages 2041--2050, 2018.

\bibitem{liu2017unsupervised}
Ming-Yu Liu, Thomas Breuel, and Jan Kautz.
\newblock Unsupervised image-to-image translation networks.
\newblock In {\em Advances in neural information processing systems}, pages
  700--708, 2017.

\bibitem{liu2020lsotb}
Qiao Liu, Xin Li, Zhenyu He, Chenglong Li, Jun Li, Zikun Zhou, Di Yuan, Jing
  Li, Kai Yang, Nana Fan, et~al.
\newblock Lsotb-tir: A large-scale high-diversity thermal infrared object
  tracking benchmark.
\newblock In {\em Proceedings of the 28th ACM International Conference on
  Multimedia}, pages 3847--3856, 2020.

\bibitem{long2018conditional}
Mingsheng Long, Zhangjie Cao, Jianmin Wang, and Michael~I Jordan.
\newblock Conditional adversarial domain adaptation.
\newblock In {\em Advances in Neural Information Processing Systems}, pages
  1640--1650, 2018.

\bibitem{long2016unsupervised}
Mingsheng Long, Han Zhu, Jianmin Wang, and Michael~I Jordan.
\newblock Unsupervised domain adaptation with residual transfer networks.
\newblock In {\em Advances in neural information processing systems}, pages
  136--144, 2016.

\bibitem{maaten2008visualizing}
Laurens van~der Maaten and Geoffrey Hinton.
\newblock Visualizing data using t-sne.
\newblock {\em Journal of machine learning research}, 9(Nov):2579--2605, 2008.

\bibitem{mccartney1976optics}
Earl~J McCartney.
\newblock Optics of the atmosphere: scattering by molecules and particles.
\newblock {\em nyjw}, 1976.

\bibitem{mueller2016benchmark}
Matthias Mueller, Neil Smith, and Bernard Ghanem.
\newblock A benchmark and simulator for uav tracking.
\newblock In {\em European conference on computer vision}, pages 445--461.
  Springer, 2016.

\bibitem{muller2018trackingnet}
Matthias Muller, Adel Bibi, Silvio Giancola, Salman Alsubaihi, and Bernard
  Ghanem.
\newblock Trackingnet: A large-scale dataset and benchmark for object tracking
  in the wild.
\newblock In {\em Proceedings of the European Conference on Computer Vision
  (ECCV)}, pages 300--317, 2018.

\bibitem{real2017youtube}
Esteban Real, Jonathon Shlens, Stefano Mazzocchi, Xin Pan, and Vincent
  Vanhoucke.
\newblock Youtube-boundingboxes: A large high-precision human-annotated data
  set for object detection in video.
\newblock In {\em Proceedings of the IEEE Conference on Computer Vision and
  Pattern Recognition}, pages 5296--5305, 2017.

\bibitem{ren2015faster}
Shaoqing Ren, Kaiming He, Ross Girshick, and Jian Sun.
\newblock Faster r-cnn: Towards real-time object detection with region proposal
  networks.
\newblock In {\em Advances in neural information processing systems}, pages
  91--99, 2015.

\bibitem{russakovsky2015imagenet}
Olga Russakovsky, Jia Deng, Hao Su, Jonathan Krause, Sanjeev Satheesh, Sean Ma,
  Zhiheng Huang, Andrej Karpathy, Aditya Khosla, Michael Bernstein, et~al.
\newblock Imagenet large scale visual recognition challenge.
\newblock {\em International journal of computer vision}, 115(3):211--252,
  2015.

\bibitem{tao2016siamese}
Ran Tao, Efstratios Gavves, and Arnold~WM Smeulders.
\newblock Siamese instance search for tracking.
\newblock In {\em Proceedings of the IEEE conference on computer vision and
  pattern recognition}, pages 1420--1429, 2016.

\bibitem{wang2019spm}
Guangting Wang, Chong Luo, Zhiwei Xiong, and Wenjun Zeng.
\newblock Spm-tracker: Series-parallel matching for real-time visual object
  tracking.
\newblock In {\em Proceedings of the IEEE Conference on Computer Vision and
  Pattern Recognition}, pages 3643--3652, 2019.

\bibitem{wang2018lstn}
Shanshan Wang and Lei Zhang.
\newblock Lstn: Latent subspace transfer network for unsupervised domain
  adaptation.
\newblock In {\em Chinese Conference on Pattern Recognition and Computer Vision
  (PRCV)}, pages 273--284. Springer, 2018.

\bibitem{7001050}
Y. {Wu}, J. {Lim}, and M. {Yang}.
\newblock Object tracking benchmark.
\newblock {\em IEEE Transactions on Pattern Analysis and Machine Intelligence},
  37(9):1834--1848, 2015.

\bibitem{zhang2017hazerd}
Yanfu Zhang, Li Ding, and Gaurav Sharma.
\newblock Hazerd: an outdoor scene dataset and benchmark for single image
  dehazing.
\newblock In {\em 2017 IEEE international conference on image processing
  (ICIP)}, pages 3205--3209. IEEE, 2017.

\bibitem{zhang2019deeper}
Zhipeng Zhang and Houwen Peng.
\newblock Deeper and wider siamese networks for real-time visual tracking.
\newblock In {\em Proceedings of the IEEE Conference on Computer Vision and
  Pattern Recognition}, pages 4591--4600, 2019.

\bibitem{zhang2020hard}
Zhongzhou Zhang and Lei Zhang.
\newblock Hard negative samples emphasis tracker without anchors.
\newblock In {\em Proceedings of the 28th ACM International Conference on
  Multimedia}, pages 4299--4308, 2020.

\end{thebibliography}
}

\end{document}